\documentclass[journal]{IEEEtran}
\usepackage{amsmath,amsfonts}
\usepackage[linesnumbered,ruled,vlined]{algorithm2e}
\usepackage{array}
\usepackage[caption=false,font=normalsize,labelfont=sf,textfont=sf]{subfig}
\usepackage{textcomp}
\usepackage{stfloats}
\usepackage{url}
\usepackage{verbatim}
\usepackage{graphicx}
\usepackage{cite}
\usepackage{breqn}
\usepackage{xcolor}
\usepackage{amsthm}
\theoremstyle{definition}
\newtheorem{definition}{Definition}[section]
\usepackage{float}
\usepackage{hyperref}
\usepackage{booktabs}
\usepackage{algpseudocode}
\usepackage{standalone}
\algnewcommand{\algorithmicand}{\textbf{ and }}
\algnewcommand{\algorithmicor}{\textbf{ or }}
\algnewcommand{\OR}{\algorithmicor}
\algnewcommand{\AND}{\algorithmicand}
\algnewcommand{\var}{\texttt}
\SetKwInput{KwInput}{Input}
\SetKwInput{KwOutput}{Output}

\SetCommentSty{mycommfont}

\begin{document}

\title{Explicit-Implicit Subgoal Planning for Long-Horizon Tasks with Sparse Reward}

\author{Fangyuan Wang, Anqing Duan, Peng Zhou, Shengzeng Huo, Guodong Guo,~\IEEEmembership{Senior~Member,~IEEE},\\Chenguang Yang,~\IEEEmembership{Fellow,~IEEE,} and David Navarro-Alarcon,~\IEEEmembership{Senior~Member,~IEEE}
    \thanks{This work is supported in part by the Research Grants Council (RGC) of Hong Kong under grant 15212721 and grant 15231023, and in part by the PolyU-EIT Collaborative PhD Training Programme under application number 220724983. \textit{Corresponding author: David Navarro-Alarcon.}}%
    \thanks{F. Wang, A. Duan, S. Huo and D. Navarro-Alarcon are with the Department of Mechanical Engineering, The Hong Kong Polytechnic University (PolyU), Kowloon, Hong Kong.
        (e-mail: fangyuan.wang@connect.polyu.hk, anqing.duan@polyu.edu.hk, kyle-sz.huo@connect.polyu.hk, dnavar@polyu.edu.hk)}
    \thanks{P. Zhou is with the Department of Computer Science, The University of Hong Kong (HKU), Pok Fu Lam, Hong Kong. (e-mail: jeffzhou@hku.hk)}
    \thanks{G. Guo is with the Ningbo Institute of Digital Twin, Eastern Institute of Technology (EIT), China. (e-mail: gdguo@eitech.edu.cn)}
    \thanks{C. Yang is with the Department of Computer Science, University of Liverpool, Liverpool, UK. (e-mail: Chenguang.Yang@liverpool.ac.uk)}
}

\maketitle

\begin{abstract}
    The challenges inherent in long-horizon tasks in robotics persist due to the typical inefficient exploration and sparse rewards in traditional reinforcement learning approaches. To address these challenges, we have developed a novel algorithm, termed Explicit-Implicit Subgoal Planning (EISP), designed to tackle long-horizon tasks through a divide-and-conquer approach. We utilize two primary criteria, feasibility and optimality, to ensure the quality of the generated subgoals. EISP consists of three components: a hybrid subgoal generator, a hindsight sampler, and a value selector. The hybrid subgoal generator uses an explicit model to infer subgoals and an implicit model to predict the final goal, inspired by way of human thinking that infers subgoals by using the current state and final goal as well as reason about the final goal conditioned on the current state and given subgoals.
    Additionally, the hindsight sampler selects valid subgoals from an offline dataset to enhance the feasibility of the generated subgoals. While the value selector utilizes the value function in reinforcement learning to filter the optimal subgoals from subgoal candidates.
    To validate our method, we conduct four long-horizon tasks in both simulation and the real world. The obtained quantitative and qualitative data indicate that our approach achieves promising performance compared to other baseline methods. These experimental results can be seen on the website \url{https://sites.google.com/view/vaesi}.
\end{abstract}

\def\abstractname{Note to Practitioners}
\begin{abstract}
    This paper addresses the persistent challenges in executing long-horizon tasks in robotics, which are hindered by inefficient exploration and sparse rewards in traditional reinforcement learning methods.
    Existing methods typically struggle with the complexity of long-horizon tasks due to their inability to effectively break down these tasks into manageable subgoals. EISP overcomes this limitation by employing a hybrid subgoal generator, a hindsight sampler, and a value selector.
    The hybrid subgoal generator utilizes an explicit model to infer subgoals and an implicit model to approximate the subgoals by maximizing the similarity between the final goal and the reconstructed goal.
    EISP has been validated through rigorous testing on four long-horizon tasks in both simulation and real-world environments.
    The results demonstrate that EISP offers superior performance compared to existing baseline methods.
    While preliminary results are promising, future research will delve into integrating with advanced exploration methods and explore its application across a broder range of tasks.
\end{abstract}


\IEEEpeerreviewmaketitle

\section{Introduction}
\IEEEPARstart{H}{uman} daily activities often involve performing long-horizon tasks, which are characterized by hierarchical structures and multiple distinct types of actions \cite{hartmann2022long, fang2022planning}. These tasks typically comprise numerous simpler yet sequential subtasks, the successful execution of which necessitates long-horizon planning and reasoning capabilities \cite{xu2018neural, sohn2020meta}.
Humans generally excel at performing long-horizon tasks, largely due to our understanding of the subgoal dependency structure inherent to the task of interest.
Although recent advances in the field of intelligent control have demonstrated the effectiveness of robots in performing complex tasks through imitation and reinforcement learning \cite{lim2022multitask, sohn2020meta, stulp2012reinforcement}, state-of-the-art methods are typically limited to completing either short-term tasks with intensive rewards or long-horizon tasks guided by a series of instructions provided by humans \cite{borras2020grasping}.
Empowering robots with the ability to reason and plan long-horizon tasks could increase the areas in which robots can assist or even replace humans, further expanding the frontiers of robotic automation.

\begin{figure}[!t]
    \centering
    \includegraphics[width=\columnwidth]{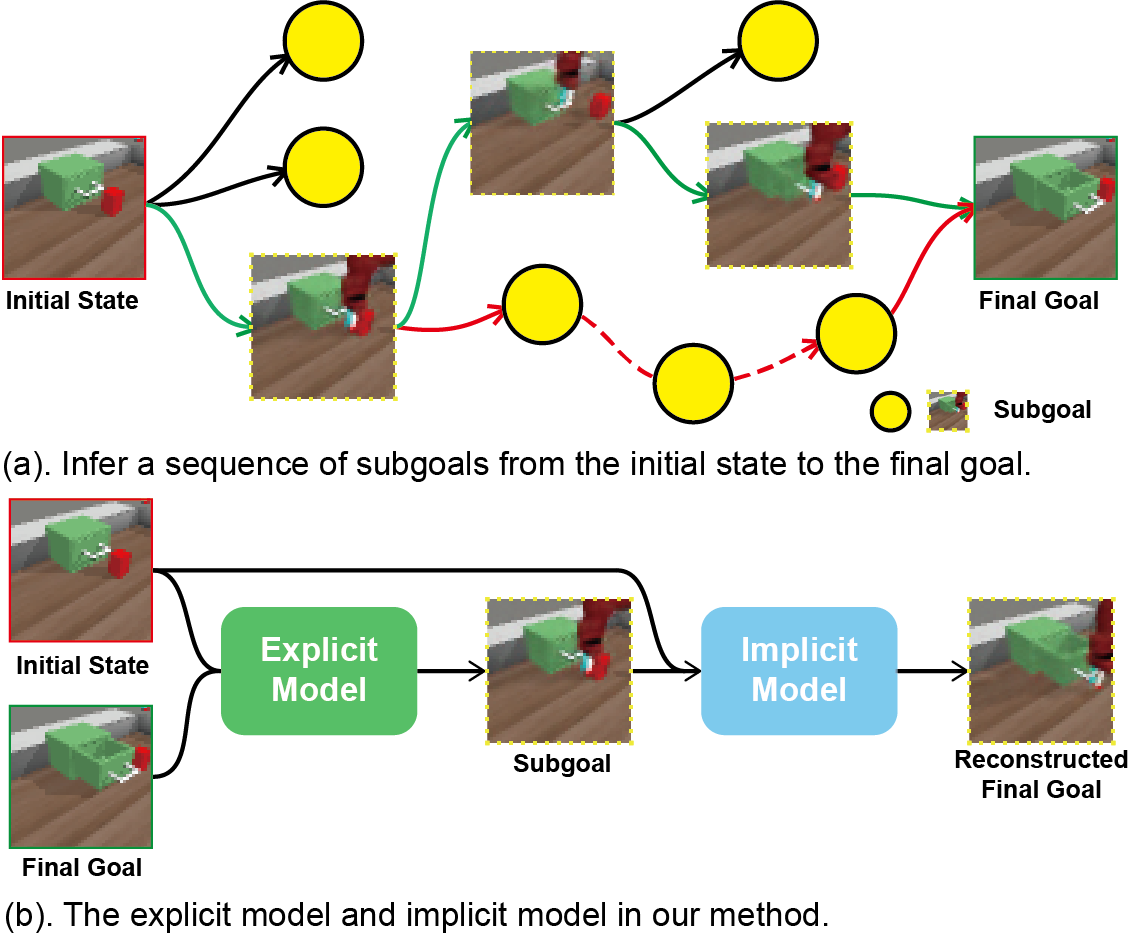}
    \caption{(a) EISP infers multiple subgoals at different task stages. Green and red lines represent possible subgoal sequences. (b) We leverage the explicit model to infer subgoals and the implicit model to predict the final goal. }
    \label{fig:insight}
\end{figure}

Current efforts in robotics to tackle long-horizon tasks involve, e.g., imitation of expert trajectories \cite{paullearning, nair2018overcoming, lim2022multitask}, enhancement of goal-oriented exploration \cite{pong2020skewfit, nairvisual, pitismaximum, florensa2018automatic, liu2022goalconditioned} and divide-and-conquer tactics \cite{lai2020hindsight, fang2022planning, nasiriany2019planning}.
For instance, imitation learning learns effective strategies to generate subgoals by extracting key points from expert demonstrations.
Yet, it usually suffers from time-consuming demonstration collection and sub-optimal expert strategies.
While enhancing the exploration efficiency of robots is essential for the discovery of effective and feasible subgoals, the exploration of goals remains limited to neighboring states of the explored region.
Given the conspicuous lack of experience with rarely explored state spaces, a significant gap remains in the efficiency requisite for more generalized reinforcement learning applications \cite{wu2022goal}.
Recently, some research has focused on decomposing complex tasks into simpler ones for solving long-horizon problems.
However, most existing methods \cite{lai2020hindsight, fang2022planning, nasiriany2019planning} cannot simultaneously guarantee the feasibility and optimality of the generated subgoal sequence and often operate under the strong assumption that the state space and the goal space are the same.
Thus, developing a robot capable of solving an extensive range of combinatorial decision problems continues to pose a long-standing challenge.

To alleviate these challenges, we propose Explicit-Implicit Subgoal Planning (EISP) that solves the long horizon problem by inferring easy-to-achieve subgoals and then accomplishing each subgoal sequentially.
Fig.\ref{fig:insight} (a) illustrates the potential subgoals that can be generated at different stages from the initial state to the final goal. The subgoal sequences connected by the green line and the red line represent possible approaches to accomplishing the long-horizon task. Because the subgoal sequence linked by the red line may require more time to complete the task, we consider the subgoal sequence connected by the green line is more optimal.
Our study mainly focuses on the subgoal generation process (without delving into the underlying subgoal approach process).
Inspired by the human approach to solving long-horizon tasks by breaking them down into smaller and manageable components, it is intuitive to employ an explicit model that takes as input the current state and final goal and outputs the subgoals, as demonstrated in previous studies \cite{fang2022planning, lai2020hindsight}. However, a critical aspect often overlooked is humans' ability to predict long-term outcomes by identifying short-term subgoals based on the current state.
As depicted in Fig.\ref{fig:insight} (b), we formulate two models: one for inferring subgoals based on the current state and the final goal, and the other for reconstructing the final goal using the current state and the given subgoals. The former model explicitly generates the subgoal, whereas the latter provides a guarantee on the worst-case for the log-likelihood of the subgoal distribution.
As suggested in \cite{florence2022implicit}, we anticipate that adding the implicit model may outperform using only an explicit model when accomplishing subgoal generation for long-horizon tasks.

The contributions of this study are outlined as follows:
\begin{enumerate}
    \item We introduce a \textbf{Hybrid Subgoal Generator}, which employs both explicit and implicit models to generate accurate subgoals by maximizing the evidence lower bound of the final goal.
    \item We quantify the quality of inferred subgoals by establishing two criteria: feasibility and optimality. We employ a \textbf{Hindsight Sampler} to facilitate the feasibility of the generated subgoals and a \textbf{Value Selector} to facilitate the selection of optimal subgoals.
    \item We conduct a series of simulations and real-world experiments to evaluate the performance of our algorithm. The results demonstrate that our method outperforms other approaches under comparable environmental conditions.
\end{enumerate}

The rest of this paper is organized as follows: Sec. \ref{sec:related} reviews the state-of-the-art; Sec. \ref{sec:preliminary} introduces the preliminaries; Sec. \ref{sec:formulation} formulates the problem; Sec. \ref{sec:method} describes the proposed methodology; Sec. \ref{sec:exp} presents the experimental results; Sec. \ref{sec:conclusion} gives conclusions and future work.

\section{Related Work}\label{sec:related}
Deep reinforcement learning (DRL) can seek an optimal solution for a given task through well-designed reward functions. Researchers have built on the recent success of DRL by extending this technique to long-horizon tasks with varying goals, known as goal-conditioned reinforcement learning (GCRL).
Current approaches to solving the long-horizon GCRL problem are mainly imitating expert trajectories, increasing the goal-orientated exploration space, and decomposing complex tasks into small ones.

\subsection{Imitation Learning}
Recent advancements \cite{nair2018overcoming, lim2022multitask} in imitation learning reveal the capacity of humans to identify and learn critical points from expert demonstrations or trajectories to accomplish the final goal. For instance, Paul et al. \cite{paullearning} derive a mapping from current states to subgoals using labels directly acquired from expert trajectories. Jin et al. \cite{jin2022learning} propose a method enabling the robot to learn an objective function from a few sparsely demonstrated keyframes to solve long-horizon tasks.
\cite{huang2023toward} learns dual-arm manipulation tasks from demonstrations generated by a teleoperation system.
Joey et al. \cite{hejna2023improving} first utilize representation learning with an instruction prediction loss that operates at a high level of abstraction to accelerate imitation learning results.
However, since the trajectories generated by human experts cannot cover the entire region of the state space, the trained policies generalize poorly to unseen tasks.
In addition, the strategies trained through imitation learning may not be optimal since the demonstration data probably is not optimal and the trained strategies are unlikely to surpass the demonstration policy.
Our approach does not require expert demonstrations and instead employs Reinforcement Learning (RL) to explore the optimal policy through trial and error.

\subsection{Exploration efficiency}
Increasing the exploration efficiency of robots is crucial for solving long-horizon tasks.
Numerous studies have endeavored to enhance exploration efficiency by enriching the experience replay buffer to incorporate more unknown states. Works such as Skew-Fit \cite{pong2020skewfit}, RIG \cite{nairvisual}, and MEGA \cite{pitismaximum} prioritize exploring low-density or sparse regions, with the intent of maximizing the likelihood of visiting fewer states, which is equivalent to maximizing the entropy of the desired goal distribution. Warde-Farly et al. \cite{warde-farley2018unsupervised}, on the other hand, aim to learn goal-conditional policies and reward functions by maximizing the mutual information between the achieved and final goals. Conversely, Hartikainen et al. \cite{hartikainen2020dynamical} chose subgoals that maximize the distance from the initial state, thereby maximizing the entropy of the desired goal distribution.
In general, these methods expand the exploration space but continue to depend on uniformly distributed actions and random action noise. Thus, goal exploration remains limited to neighboring states near the explored region.
Our work tackles the exploration challenge by automatically generating subgoals that can be predicted on the way to the long-horizon final goal.

\subsection{Subgoal orientated goal-conditioned RL}
Drawing inspiration from the human tendency to decompose long-horizon tasks into smaller ones, some researchers are exploring the generation of subgoals by selecting them from historical experiences \cite{chane-sane2021goalconditioned, lai2020hindsight, eysenbachsearch} or by training subgoal models \cite{fang2022planning, nasiriany2019planning}. The experience replay buffer, rich in valuable historical data, facilitates the extraction of valid states as subgoals in a simple and efficient manner, while also ensuring their reachability.
However, some methods necessitate the explicit specification of the number of subgoals or decision layers and rely solely on the initial state to predict future subgoals \cite{fang2022planning}.
And distance-based methods, like \cite{fang2022planning, nasiriany2019planning} strive to maximize the distance between the subgoals and the initial state to obtain better subgoals. They typically assume the co-existence of states and goals within the same space. This assumption does not hold true in numerous environments. For example, in the Stack environment, the state comprises the information of blocks and gripper, whereas the goal pertains solely to the goal position of blocks.

Recently, new image editing diffusion models like SuSIE \cite{black2023zero} and SkillDiffuser \cite{liang2023skilldiffuser} have been introduced for subgoal generation. Despite their powerful capabilities, these models depend on pretrained large diffusion models and robot demonstration video data, and are confined to image-based environments. Moreover, the substantial computational resources required for these large models pose a challenge for deployment in real-world robotic tasks.


Contrary to previous work that infers subgoals exclusively using an explicit model, we restrict our policy search by employing an implicit model and the prior distribution implicitly represented in the offline dataset.
Our work also adopts a divide-and-conquer manner to solve long-horizon tasks by utilizing the hierarchical architecture \cite{dayan1992feudal, dietterich2000hierarchical, levy2017learning}, where the top-level focuses on reasoning and decision-making to generate subgoals, and the low-level interacts with the environment to achieve the subgoals sequentially.

\section{Preliminaries}\label{sec:preliminary}
\subsection{Semi-MDP}
We formulate the long-horizon task as a Semi-Markov Decision Process (Semi-MDP) \cite{sutton1999mdps}. More precisely, we extend the general MDP by adding option policy $\psi$ and options $\mathcal{O}$. The Semi-MDP is denoted by $\langle \mathcal{S}, \mathcal{A}, r, \rho_0, \rho_g, \mathcal{T}, \gamma, \mathcal{G}, \mathcal{O}, \psi \rangle$, where $\mathcal{S}$ is state space containing all possible state, $\mathcal{A}$ is action space and $r$ is the reward function. We use $\rho_0$ and $\rho_g$ to denote the distribution of initial state and final goal, respectively. $\mathcal{T}: \mathcal{S}\times \mathcal{A}\rightarrow \mathcal{S}$ is the transition function and $\gamma$ is the discount factor. $\mathcal{G}$ is the goal space from which final goals and subgoals can be sampled.
$\mathcal{O}: \langle \mathcal{I}, \pi, \mathcal{E} \rangle$ is the option consisting of three components, where $\mathcal{I} \in \mathcal{S}$ is the initial state space, $\mathcal{E} \in \mathcal{S}$ is the terminal state space, $\pi$ is the specific policy for current option, respectively.

We denote the achieved goal at time step $t$ as $ag_t$. Typically, $ag_t = \phi(s_t)$, where $\phi: \mathcal{S} \rightarrow \mathcal{G}$ represents a known and traceable mapping that defines goal representation \cite{renexploration}. The subgoal is often identified as the achieved goal at the terminal state of an option.
$\psi: \mathcal{S} \times \mathcal{G} \rightarrow \mathcal{G}$ is the option policy mapping from state and final goal to subgoal for inferring a sequential option $\{o_1,o_2,\cdots,o_n \}$ where each $o_i \in \mathcal{O}$.
And for each option $o_i$, the policy $\pi_{i}$ is utilized when the agent encounters the initial state $s_0^i \in \mathcal{I}$ and terminates in $s_e^i \in \mathcal{E}$.

In sparse reward goal-conditioned RL, the reward function $r$ is a binary signal indicating whether the current goal is achieved, as expressed by,
\begin{equation}
    r(s_t, a_t, g) =
    \begin{cases}
        0  & ||\phi(s_t) - g||_2 \le \epsilon, \\
        -1 & \text{otherwise}
    \end{cases}
\end{equation}

\noindent where $\epsilon$ is a small threshold indicating whether the goal is considered to be reached, and $s_t$, $a_t$ are the state and action at time step $t$ and $g$ is the final goal.
Denote that the initial state $s_0$ of the whole trajectory is the initial state of option $o_1$, and the final state is the terminal state of option $o_n$. And the terminal state $s_e^i$ of option $o_i$ is the initial state $s^{i+1}_0$ of next option $o_{i+1}$. All options are connected in a head-to-tail manner to form the entire trajectory.

\subsection{Soft Actor-Critic}
Soft Actor-Critic (SAC) \cite{haarnoja2018soft} is an off-policy reinforcement learning algorithm that attempts to maximize the expected return and increase the policy's entropy. The underlying idea is that increasing entropy leads to more exploration and effectively prevents the policy from reaching a local optimum.

Let $\tau = \{s_0, a_0, s_1, a_1, \cdots\}$ be the trajectory starting from initial state $s_0$. We use $\rho_{\tau} = \rho_0 \prod_t \pi(a_t | s_t, g) \mathcal{T}(s_{t+1}|s_t, a_t)$ to denote the distribution of the $\tau$ induced by the policy $\pi$.


In contrast to the standard maximum expected reward used in traditional reinforcement learning, the objective of action policy $\pi$ aims to maximize its entropy at each visited state by augmenting with an entropy term,

\begin{dmath}
    J(\pi) = \mathbb{E}_{g\sim\rho_g, \tau\sim \rho_{\tau}} \left[ \sum_{t} \gamma^t r(s_t, a_t, g) + \alpha \mathcal{H}(\pi(\cdot|s_t,g)) \right]
\end{dmath}

\noindent where $\mathcal{H}(\pi(s_t, g))$ is the entropy of the policy $\pi$, and $\alpha > 0$ is the temperature parameter that determines the importance of the entropy term with respect to the reward and is used to control the stochasticity of the optimal policy.

Thus, we can obtain the optimal action policy $\pi^*$ by:

\begin{dmath} \label{sac_objective}
    \pi^* = \arg\max_{\pi} J
\end{dmath}

\subsection{Hindsight Experience Replay}
Hindsight Experience Replay (HER) \cite{andrychowicz2017hindsight} is an algorithm that improves data efficiency by relabeling data in the replay buffer. It builds on the basis that failed trajectories may still achieve other unexpected states, which could be helpful for learning a policy.

It mitigates the challenge of sparse reward settings by constructing imagined goals in a simple heuristic way to leverage previous replays. During the training process, the HER relabeled the desired goal of this trajectory by specific sampling strategies, such as \texttt{future} and \texttt{final}, which take the future achieved goal and the final achieved goal as the desired goal, respectively.

\section{Problem Formulation}\label{sec:formulation}
Given an initial state $s_0 \in S$ and desired final goal $g \in \mathcal{G}$, our objective is to infer a sequence of feasible and optimal options $\{o_1,o_2,\cdots,o_n \}$ by using option policy $\psi$, where $o_i = \langle s_0^i, \pi_i, s_e^i \rangle, i \in [1,n]$.
As the terminating state $s_e^i$ becomes the initial state for the subsequent option, the option inference problem can be simplified to a terminal state inference problem. In other words, we aim to infer a sequence of feasible and optimal subgoals $\{\hat{g}_1, \hat{g}_2, \cdots, \hat{g}_{n-1} \}$, where $\hat{g}_{i} = \phi(s_e^i)$ for $i = [1, n-1]$.

\begin{definition}[Feasibility]
    The sequence of subgoals $\hat{g}_1, \hat{g}_2, \cdots, \hat{g}_{n-1}$ is feasible if the robot can approach the subgoal $\hat{g}_{i}$ from the previous one $\hat{g}_{i-1}$.
\end{definition}

Within the subgoal space $\mathcal{G}$, the majority of subgoals generated by the option policy are unfeasible, either due to spatial distance or the strict dependency order of subtasks inherent in long-horizon tasks \cite{sohnhierarchicala}.
An option policy $\psi$ is anticipated to generate a subgoal sequence $\hat{g}_1, \hat{g}_2, \cdots, \hat{g}_{n-1}$, such that the transition probability from previous subgoal $\hat{g}_{i-1}$ to current subgoal $\hat{g}_i$ exceeds 0. Here, the transition probability is denoted as $\rho(s_0^i)\prod_{s = s_{0}^i, a \sim \pi_i}^{s_e^i} \left[ \pi_{i}(s, \hat{g}_i)\mathcal{T}(s, a) \right]$ and $\rho(s_0^i)$ is the probability of state $s_0^i$. More details can be found in Appendix \ref{sec:feasibility}.

\begin{definition}[Optimality]
    We call a sequence of subgoals $\hat{g}_1^*, \hat{g}_2^*, \cdots, \hat{g}_{n-1}^*$ optimal only if the robot can achieve the maximum accumulated reward of all options to finish the task.
\end{definition}

Resolving long-horizon tasks can be distilled down to the challenge of optimization over a sequence of subgoals for the goal-conditioned policy. This optimization over subgoals can be viewed as high-level planning, wherein the optimizer identifies waypoints to accomplish each subgoal \cite{fang2022planning}. We aim to find an optimal option policy $\psi^*$ to maximize the total accumulated reward of all options,

\begin{equation}
    \label{}
    \psi^{*} = \arg\max_{\psi} \sum_{i=1}^{n} J_{o_i}
\end{equation}

\noindent where $J_{o_i}$ is the objective of option $o_i$, denoted as

\begin{equation}
    J_{o_i}(\psi, \pi_i) = \mathbb{E}_{\hat{g}_i \sim \psi, a_t \sim \pi_i} \Big[ \sum_{t} \gamma^t r(s_t^i, a_t, \hat{g}_i) + \alpha \mathcal{H}(\pi(\cdot|s_t^i, \hat{g})) \Big]
\end{equation}

\begin{figure}[t]
    \centering
    \includegraphics[width=\columnwidth]{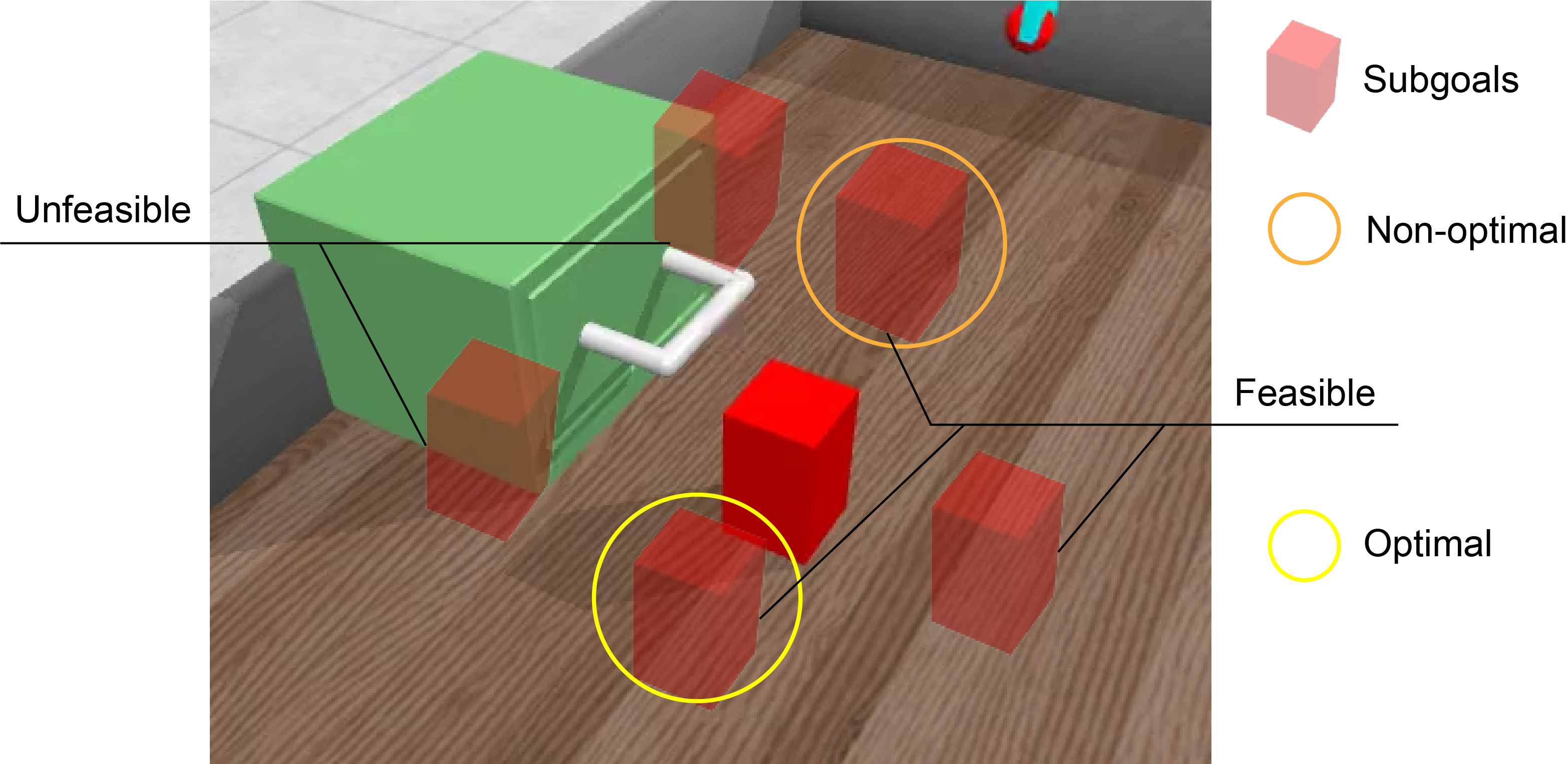}
    \caption{Examples of feasible and optimal subgoals in the OpenDrawer environment are depicted.
    }
    \label{fig:require}
\end{figure}

We employ the OpenDrawer environment as a case study to elucidate the concept of feasible and optimal subgoals. As illustrated in Fig. \ref{fig:require}, the ultimate objective is to open the drawer by manipulating the handle. However, a red obstacle block is positioned in front of the drawer, necessitating that the robotic arm first push the block away. This task encompasses two distinct skills: pushing the block and opening the drawer. The option policy infers subgoals within the subgoal space, which can sometimes interfere with the drawer, rendering them unfeasible for the robot. Considering the spatial relationship between the block and the drawer, the subgoal encircled by the yellow line is deemed optimal as it enables the robot to move the block away more efficiently.


\section{Methodology}\label{sec:method}
\begin{figure*}[!ht]
    \centering
    \includegraphics[width=\textwidth]{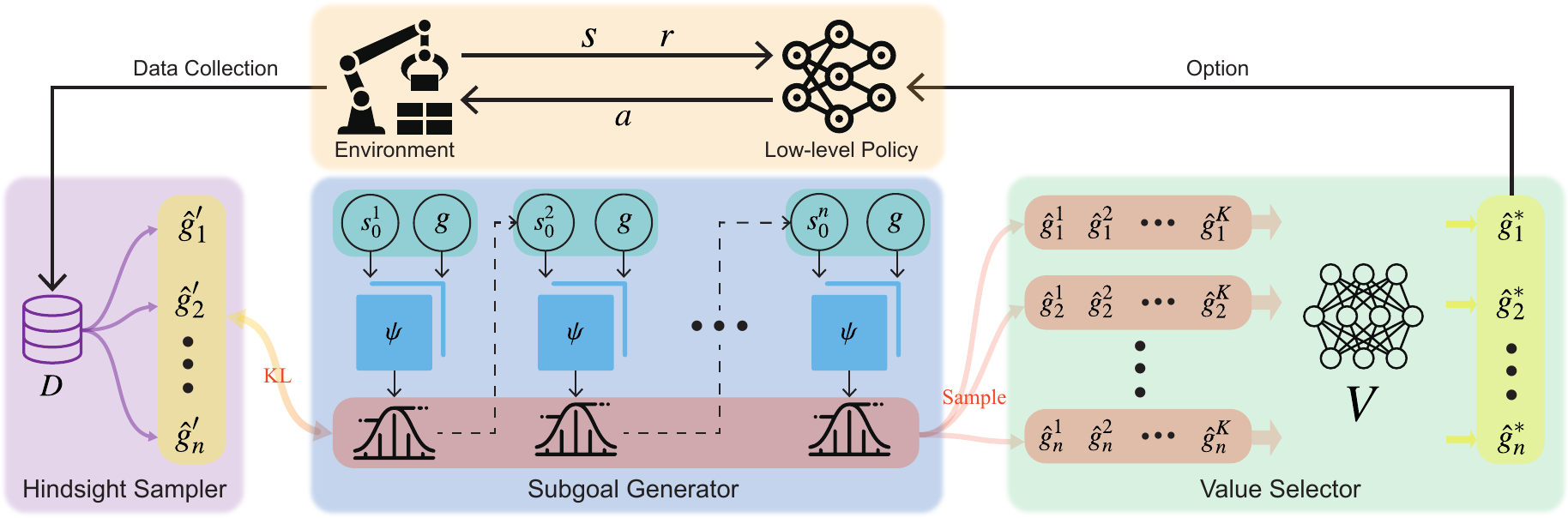}
    \caption{The Explicit-Implicit Subgoal Planning (EISP) algorithm consists of three main components: a Hybrid Subgoal Generator, a Hindsight Sampler, and a Value Selector. The subgoal generator takes as input the current state and the desired goal and outputs subgoals to accomplish the long-horizon task. The Value Selector and the Hindsight Sampler are utilized to ensure that the subgoals are optimal and feasible, respectively.}
    \label{fig:algo}
\end{figure*}

As illustrated in Fig. \ref{fig:algo}, our EISP algorithm comprises three components: the Hybrid Subgoal Generator, the Hindsight Sampler, and the Value Selector.

\subsection{Hybrid Subgoal Generator}
Given an initial state $s_0$ and a final goal $g$, we expect the robot to infer a series of subgoals $\{\hat{g}_1, \hat{g}_2, \cdots, \hat{g}_{n-1}\}$ to guide it to the final goal $g$.
These subgoals are subsequently utilized to bootstrap the low-level action policy $\pi$, thereby generating the actions to accomplish each subgoal.

Generally, most current subgoal generators \cite{fang2022planning, chane-sane2021goalconditioned, lai2020hindsight} employ an explicit feed-forward model that maps subgoals directly from the state and the final goal. These methods are inspired by the human approach to solving long-horizon tasks, wherein humans generate subgoals to decompose long-horizon tasks into smaller, more manageable ones based on the final goal and the current state. However, one critical aspect often overlooked is that humans can also predict long-term outcomes using short-term subgoals conditioned on the current state. Current research \cite{florence2022implicit} in behavior cloning has demonstrated that implicit models possess a superior capacity to learn long-horizon tasks more effectively than their explicit counterparts. This evidence incites us to employ the implicit model in subgoal generators, which is anticipated to enhance the ability to solve long-horizon tasks significantly.

\begin{figure}[t]
    \centering
    \includegraphics[width=\columnwidth]{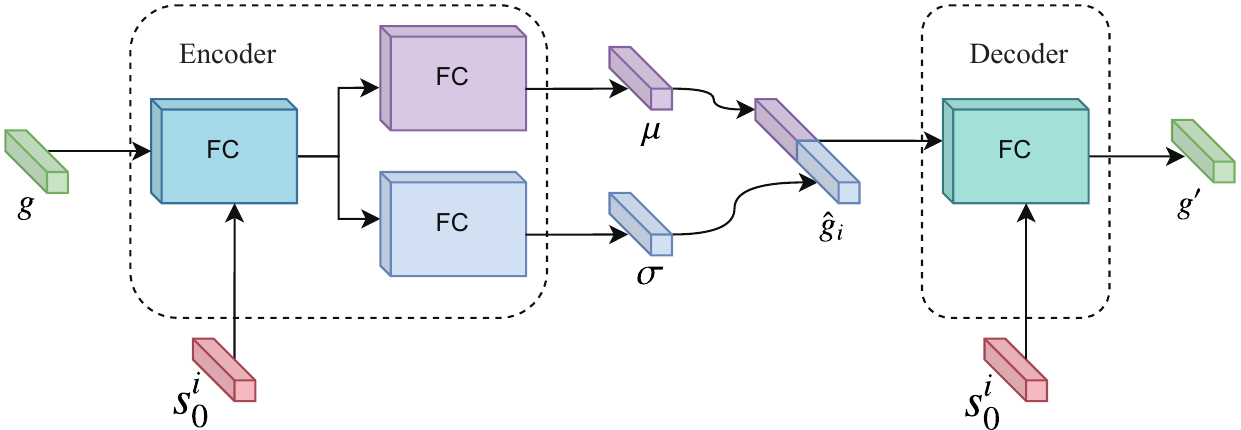}
    \caption{Details of our proposed subgoal inference method. It inherits the variational autoencoder structure, where the encoder generates the subgoals by taking the current state and the final goal as inputs, and the decoder generates a reconstructed final goal conditioned on the current state and the subgoal.}
    \label{fig:vae}
\end{figure}

Based on these insights, we propose a hybrid subgoal generator, fundamentally structured as a Conditional Variational Autoencoder (CVAE) \cite{sohnlearning}. Within this framework, the encoder $\psi$ serves as an explicit model that can infer a sequence of subgoals based on the current state and the final goal, while the decoder $\psi^{\prime}$ functions as an implicit model, designed to reconstruct the final goal using the current state and the given subgoals, as depicted in Fig. \ref{fig:vae}.
After preprocessing the state $s$ and the final goal $g$ through several fully connected layers, the encoder computes the mean $\mu$ and the standard deviation $\sigma$ via two fully connected networks. The subgoal is then sampled from the normal distribution modeled using these calculated values of $\mu$ and $\sigma$. The decoder takes the subgoal $\hat{g}_i$ as input and conditioned on the current state $s_0^i$ to generate the reconstructed final goal $g^{\prime}$. Notably, we also employ the reparameterization trick to make the encoder and decoder differentiable.

Unlike the original CVAE, where the latent space serves as a low-dimensional mapping from the inputs, we employ the CVAE structure but interpret the latent space as the subgoal space. This approach distinguishes our work from previous studies \cite{fang2022planning}, which also use the CVAE structure to generate subgoals but focus on reconstructing the subgoal using a learned latent representation of the transitions.
By modeling the conditional option policy $\psi$, which takes as inputs the initial state $s_0^i$ of option $i$ and the final goal $g$, we transition the subgoal generation approach from the initial planner (which infers subgoals only at the initial state) to an incremental Planner (which infers subgoals based on current state).
That is, the subgoals can be inferred incrementally by:

\begin{equation}
    \label{}
    \hat{g}_i = \psi(s_0^i,g), \quad \text{for } i = 1,\ldots, n-1
\end{equation}

\noindent where $s_0^1 = s_0$.
To ensure the robustness of the inferred subgoals and prevent the agent from lingering at one subgoal, we impose a time limit of $T$. The option policy $\psi$ deduces a new subgoal either when the agent reaches the current subgoal or when the allocated time step for the current subgoal expires. This can be expressed as follows:
\begin{equation}
    \label{}
    s_0^i =
    \begin{cases}
        s^{i-1}_e & \text{reach the subgoal $\hat{g}_{i-1}$ in $T$ time steps}, \\
        s^{i-1}_T & \text{otherwise}
    \end{cases}
\end{equation}
\noindent where $s^{i-1}_e$ is the terminal state when the robot reaches the subgoal $\hat{g}_{i-1}$ of option $i-1$.
We denote the implicit model as $\psi^{\prime}(g^{\prime}|s_0^i, \hat{g}_i)$, which is designed to accurately reconstruct the final goal from the subgoal.
The objective is to maximize the log probability of the final goal, $\log{p_{\psi^{\prime}}(g)}$, by maximizing its evidence lower bound (ELBO) \cite{kingma2013auto}, formulated as:
\begin{align}
    \label{elbo}
    \mathbb{E}_{p_{\psi}(\hat{g}|g,s_0^i)}\Big[\log\frac{p_{\psi^{\prime}}(g,\hat{g}, s_0^i)}{p_{\psi}(\hat{g}|g,s_0^i)}\Big]
\end{align}
Thus, the objective of the hybrid subgoal generator $\mathcal{L}_{HB}$ can be defined as follows:
\begin{align}
    \label{vae_loss}
    \mathbb{E}_{p_{\psi}(\hat{g}_i|g,s_0^i)} \big[ \log p_{\psi^{\prime}} (g | \hat{g}_i, s_0^i) \big] - D_{KL}\big[p_{\psi}(\hat{g}_i|s_0^i, g)\|p(\hat{g}_i)\big]
\end{align}
where $p(\hat{g}_i)$ is the prior distribution over the subgoals, typically a standard normal distribution $\mathcal{N}(0, I)$. $D_{KL}[\psi(\hat{g}|s_0^i, g_i)\|p(\hat{g}_i)]$ denotes the Kullback-Leibler (KL) Divergence \cite{kingma2013auto} between the prior distribution $p(\hat{g}_i)$ and the distribution of the generated subgoals by $\psi$. A detailed derivation can be found in Appendix \ref{sec:derivation}.
We will introduce an improved prior distribution for the subgoals in the following section.

\subsection{Hindsight Sampler for Feasible Subgoals}
\begin{figure}[t]
    \centering
    \includegraphics[width=\columnwidth]{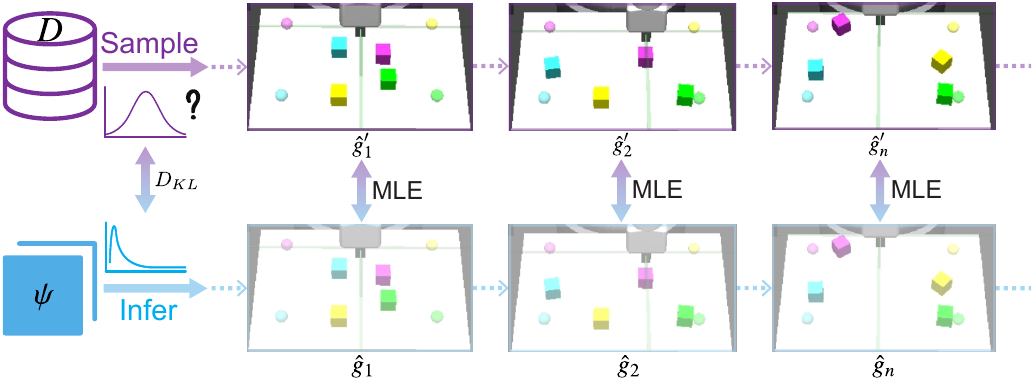}
    \caption{Hindsight Sampler to guide the generator to produce feasible subgoals.}
    \label{fig:hs}
\end{figure}

The subgoals sampled from the ground truth subgoal distribution ensure that the transition probability from the current subgoal $\hat{g}_{i-1}$ to the next subgoal $\hat{g}_{i}$ is greater than zero. Thus, the distribution of the generated subgoals should closely approximate the true subgoal distribution, which can be achieved by minimizing the KL divergence between the distribution of the generated subgoals by $\psi$ and the true valid subgoal distribution $p_{t}(\hat{g}_i)$.
However, the primary challenge arises from the unknown ground truth subgoal distribution $p_t(\hat{g}_i)$, which complicates the estimation of the KL divergence.

Empirically, the subgoal distribution within the offline dataset $\mathcal{D}$ can be approximately regarded as the ground truth subgoal distribution. Therefore, we adopt the subgoal distribution of the offline dataset as our prior distribution and aim to minimize the KL divergence between them.
To this end, we present a Hindsight Sampler, an approach based on the Hindsight Experience Replay buffer \cite{andrychowicz2017hindsight}, designed to sample valid subgoals from an offline dataset. The core idea is to regard the achieved goal reached at the end of the trajectory as the final goal and recalculate the rewards for all states in the current trajectory. This revised trajectory is then employed to train the option policy.
As shown in Fig. \ref{fig:hs}, during the training period, we first sample the trajectory $\tau_{D}$ with total time step $T_{\tau}$ from the offline dataset $\mathcal{D}$, and take the final achieved goal of this trajectory as the desired goal $g$ of the current trajectory. We then select waypoints from the trajectory as subgoals at intervals of $T_{\tau}/n$, denoted as $\left\{ \hat{g}_1^{\prime}, \hat{g}_2^{\prime}, \cdots, \hat{g}_{n-1}^{\prime} \right\}$.

We utilize the maximum likelihood estimation (MLE) \cite{rossi2018mathematical}, which is a method estimating the parameters of an assumed probability distribution given some observed data, to maximize the log-likelihood of the valid subgoals sampled from the offline dataset under the unknown option policy $\psi$. The objective of the Hindsight Sampler $L_{HS}$ is as follows:

\begin{equation}
    \label{hs_loss}
    \mathcal{L}_{HS} = -\log (p_{\psi}(\hat{g}_{i}^{\prime}))
\end{equation}

\noindent where $p_{\psi}(\hat{g}_{i}^{\prime})$ is the probability of sampled subgoals under the distribution of subgoals generated by the hybrid subgoal generator.

\begin{algorithm}[t]
    \captionsetup{labelfont={sc,bf}, labelsep=newline}
    \caption{Collect Rollout Trajectories}
    \label{algo:collect}
    \KwInput{Option policy $\psi$, low-level policy $\pi$, state value function $V$, time limit $T$, distance function $Dist$, initial state $s_0$, final goal $g$}
    \KwOutput{$\mathcal{D}$}

    $s_0 \gets$ initial state \\
    $\hat{g}^* \gets g$ \\
    \ForAll{$t \gets [0, 1, 2, \cdots]$}{
        $ag_t \gets \phi(s_t)$ \\
        \tcc{Generate and select optimal subgoal}
        \If{$t \mod T = 0$ \OR $D(ag_t, \hat{g}^*) \leq \epsilon $}{
            Sample $K$ subgoal candidates $\{\hat{g}^1, \hat{g}^2, \cdots, \hat{g}^K\}$ by using $\psi$ \\
            Select optimal subgoal $\hat{g}^*$ from candidates by using \eqref{eq:value_selector} \\
            \tcc{Set subgoal as the final goal if they are too close}
            \If{$Dist(\hat{g}^*, g) \leq \epsilon$}{
                $\hat{g}^{*} \gets g$
            }
        }
        $a_t \gets \pi(s_t, \hat{g}^*)$ \\
        Execute action $a_t$ and obtain next state $s_{t+1}$ and reward $r_t$ \\
        Store $(s_t, a_t, r_t, s_{t+1}, \hat{g}^*)$ in experience replay buffer $\mathcal{D}$ \\
    }
\end{algorithm}

\subsection{Value Selector for Optimal Subgoals}
The Universal Value Function Approximator (UVFA) plays a critical role in goal-conditioned RL. It functions as a mapping from the goal-conditioned state space to a nonlinear value function approximator \cite{schaul2015universal}. We denote the state value function as $V$. In the context of goal-conditioned RL, it is typically utilized to assess the current state under the condition of the goal.
The value function $V$ of SAC in goal-conditioned RL involves the extra entropy bonuses from every time step utilized to assess the state, denoted as follows:

\begin{equation}
    V(s, g) = \mathbb{E}_{g \sim \rho_g, \tau \sim \rho_{\tau}} \Big[ \sum_{t} \gamma^t r(s_t, a_t, g) + \alpha \mathcal{H}(\pi(\cdot|s_t, g)) \Big]
\end{equation}
where $s_0=s$.
In our algorithm, $V$ serves a dual purpose: guides the updates of low-level policy and selects optimal subgoals for the high-level planner.
To generate subgoal $\hat{g}$ at current state $s_t$, conditioned on the final goal $g$, we initially sample $K$ candidates $\left\{\hat{g}_{1}, \hat{g}^{2}, \cdots, \hat{g}^{k}, \cdots, \hat{g}^{K} \right\}$ using the Hybrid Subgoal Generator, where $\hat{g}^{k} \sim \psi(s_t, g)$.
As shown in Fig. \ref{fig:algo}, the candidates are then ranked using the state value function $V$, where higher state values represent better subgoals. Hence, the subgoal state corresponding to the highest value will be selected as the optimal subgoal $\hat{g}^{*}$ for the action policy, that is,

\begin{equation}\label{eq:value_selector}
    \hat{g}^{*} = \arg\max_{\hat{g}^{k}} \{V(s, \hat{g}^{k}), k \in [1, K]\}
\end{equation}

The optimal subgoal $\hat{g}^{*}$ is subsequently utilized to guide the low-level action policy. The value function may not be optimal during the initial stages of training as the low-level policy may not be sufficiently trained. However, as the training of the value function progresses and improves, the selected subgoals also tend towards optimality.


\subsection{Integrate with RL}
\begin{algorithm}[t]
    \captionsetup{labelfont={sc,bf}, labelsep=newline}
    \caption{EISP}
    \label{algo:EISP}
    \KwInput{$\mathcal{D}$}
    \KwOutput{$\psi^{*}, \psi^{\prime *}$}
    Sample mini-batch $\mathcal{B} \gets \{(s_t, a_t, r_t, s_{t+1}, \hat{g}^*)|^{N}_{t=1} \sim \mathcal{D} \}$ \\
    Sample subgoals $\hat{g}^{\prime}$ by using \textbf{Hindsight Sampler} \\
    Calculate $\mathcal{L}_{HY} \gets$ \eqref{vae_loss}, $\mathcal{L}_{HS} \gets$ \eqref{hs_loss} \\
    Update option policy by $\psi^*, \psi^{\prime *} = \arg\min_{\psi, \psi^{\prime}} \mathcal{L}$ \\
    Update low-level policy RL algorithms, e.g., SAC \\
\end{algorithm}

Alg. \ref{algo:collect} offers a detailed description of the process of collecting rollout experience for the experience replay buffer.
The function $D(g_1, g_2)$ is employed to estimate the distance between goals $g_1$ and $g_2$.
We will initially infer a set of potential subgoal candidates via the Hybrid Subgoal Generator when the current subgoal is achieved or the time limit $T$ is expirated. Then, the optimal subgoal is selected using the Value Selector. By replacing the final goal with the selected subgoal continuously, we can obtain trajectories of transitions $(s_t, a_t, r_t, s_{t+1}, \hat{g}^*)$ and store it to the replay buffer $\mathcal{D}$.

The entire training process of EISP is shown in Alg. \ref{algo:EISP}.
At first, a mini-batch $\mathcal{B}$ is sampled from $\mathcal{D}$, consisting of $N-1$ transitions of $(s_t, a_t, r_t, s_{t+1}, g)$, where $t \in [1, N]$.
For each transition, the final goal $g$ is relabeled by the future achieved goal, and valid subgoals $\hat{g}^{\prime}$ will be sampled by using the Hindsight Sampler. Then, the objective of the EISP is formulated as:

\begin{equation}\label{eq:loss}
    \mathcal{L} = \mathcal{L}_{HY} + \beta \mathcal{L}_{HS}
\end{equation}

\noindent where $\beta$ is the hyper parameter. The explicit policy $\psi$ and implicit policy $\psi^{\prime}$ can be updated by using stochastic gradient descent with $\mathcal{L}$. By alternating between the collection of rollout experience using Alg. \ref{algo:collect} and optimization of parameters via Alg. \ref{algo:EISP}, an optimal subgoal generator can be obtained.

Since the off-policy RL algorithms are used to train the low-level action policy, we collect trajectories by using scripted policies to pre-train the subgoal generator and then fine-tune it during the low-level training process.
The proposed subgoal inference algorithm is entirely compatible with other off-policy algorithms, owing to its exclusive focus on high-level planning. We employ the SAC method to train the low-level policy.

\section{Experiments and Analysis}
\label{sec:exp}

\begin{figure}[h]
    \centering
    \includegraphics[width=\columnwidth]{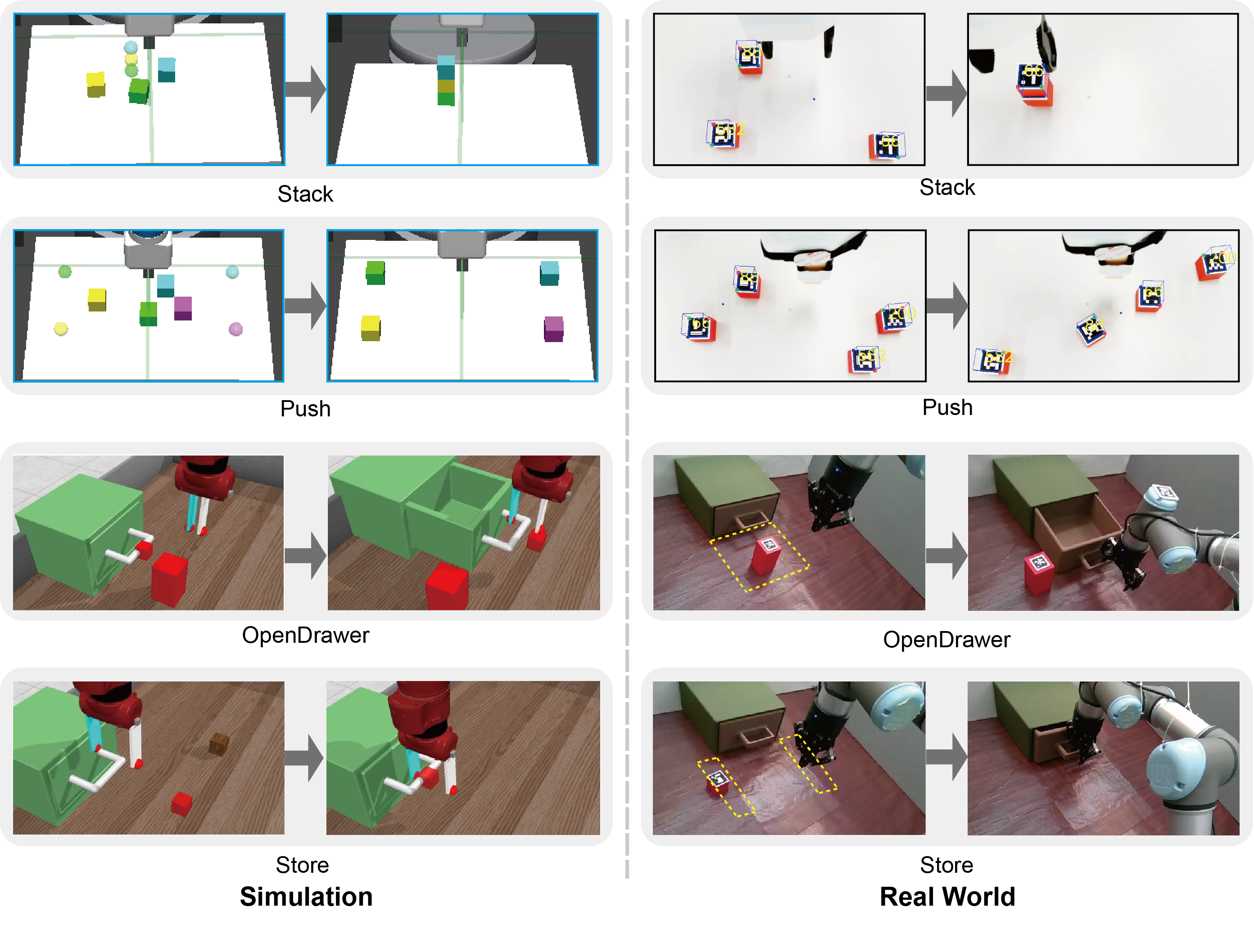}
    \caption{Four manipulation tasks, i.e., Stack, Push, OpenDrawer, and Store. For the Stack and Push tasks, we label the initial states and the desired goals as solid cubes and transparent spheres. For the OpenDrawer and Store tasks, we use yellow dashed rectangles to represent the initial position of the blocks.}
    \label{fig:env}
\end{figure}

\subsection{Tasks}
We evaluate EISP in four tasks: 1) Stack, 2) Push, 3) OpenDrawer, and 4) Store, as depicted in Fig. \ref{fig:env}. The left and right columns present the simulation tasks and real-world tasks. All simulation environments are created using the Gymnasium Robotics \cite{gymnasium2023github} Framework.
The dimensional information of the state space $\mathcal{S}$, action space $\mathcal{A}$, goal space $\mathcal{G}$, and the episode length $L$ are provided in Table \ref{tab:tb_env}.

\noindent\textbf{Stack} The block stacking task consists of picking up three blocks and placing them in sequence at the target location. The dimension of the observation space is 55, containing information about the fixture blocks. The action space contains the position of the fixture end-effector and the grasping signal. Multi-step planning makes it more difficult because we have to complete the whole task without disturbing all the completed subtasks \cite {nair2018overcoming}.

\noindent\textbf{Push4} This task combines the four block-pushing tasks of D4RL \cite{fu2020d4rl}. The observation space is 70-dimensional and consists of information about fixtures and blocks. The action space is the position of its end-effector and the gripper status. The robot must sequentially push four objects to the four target positions on the table. The situations can become more complex when some blocks are blocked by others, necessitating the robot to push the blocks in a specific order.

\noindent\textbf{OpenDrawer} The robot needs to open the drawer by first removing the block and then pulling the handle. We evaluate EISP in both the state-based and image-based environment. In the state-based environment, the observation space is 42-dimensional, containing information about the drawer, the handle, the block and the gripper. In the image-based environment, the observation space is a $3 \times 48 \times 48$ image. The action space is the position of the end-effector and the gripper status. The task is challenging due to it contains multiple skills (Pushing the block and pulling the handle) and task dependencies.

\noindent\textbf{Store} This task requires the robot to first open the drawer, then pick and place the block into the drawer, and finally close the drawer. Multiple skills like pushing, picking, placing and grasping are involved in this task. Like the OpenDrawer task, we also conduct experiments in both the state-based and image-based environment. The settings of the observation and action space are the same as the OpenDrawer task.

\begin{table}[!t]
    \centering
    \caption{Parameters setting.}
    \begin{tabular}{ccccc}
        \toprule
        Env           & Stack          & Push4          & \begin{tabular}{@{}c@{}}OpenDrawer \\ (State)\end{tabular} & \begin{tabular}{@{}c@{}}Store \\ (State)\end{tabular} \\
        \midrule
        $\mathcal{S}$ & 55             & 70             & 42                                                         & 42                                                    \\
        $\mathcal{G}$ & 9              & 12             & 6                                                          & 6                                                     \\
        $\mathcal{A}$ & 4              & 4              & 4                                                          & 4                                                     \\
        $L$           & 300            & 400            & 300                                                        & 1000                                                  \\
        \midrule
        $\mathcal{D}$ & $1e5$          & $1e6$          & $1e6$                                                      & $1e6$                                                 \\
        $N$           & $1024$         & $1024$         & $1024$                                                     & $1024$                                                \\
        SAC $lr$      & $3e-3$         & $3e-4$         & $3e-4$                                                     & $3e-3$                                                \\
        $\pi$         & $256 \times 2$ & $256 \times 3$ & $512 \times 4$                                             & $512 \times 4$                                        \\
        $\alpha$      & 0.01           & 0.01           & 0.01                                                       & 0.01                                                  \\
        $\gamma$      & $0.99$         & $0.99$         & $0.99$                                                     & $0.99$                                                \\
        \midrule
        EISP $lr$     & $1e-5$         & $1e-5$         & $1e-5$                                                     & $1e-5$                                                \\
        $T$           & 30             & 30             & 50                                                         & 50                                                    \\
        $n$           & 4              & 4              & 6                                                          & 6                                                     \\
        $\psi$        & $256 \times 2$ & $256 \times 2$ & $512 \times 4$                                             & $512 \times 4$                                        \\
        $\beta$       & $1e-2$         & $1e-2$         & $1e-3$                                                     & $1e-3$                                                \\
        \bottomrule
    \end{tabular}
    \label{tab:tb_env}
\end{table}

\subsection{Implementation Details}
The Hybrid Subgoal Generator is based on the CVAE structure and consists of an encoder $\psi$ and a decoder $\psi^{\prime}$.
The encoder takes the current state and the desired goal as inputs and uses two probabilistic neural networks to output the mean $\mu$ and deviation $\sigma$. We then parameterize the Laplace distribution with the above outputs.
We implement two probabilistic neural networks using one layer of fully connected network mapping from the encoder output to the subgoal space.
The decoder has the opposite structure to the encoder. It takes as inputs the encoder-generated subgoals and the current state and outputs the reconstructed desired goal.
The SAC is used as the underlying RL algorithm for action policy training and then fine-tuned online. We use a fixed temperature parameter $\alpha = 0.01$ for all tasks.
Table \ref{tab:tb_env} presents the specifications of various parameters including the size of the replay buffer $\mathcal{D}$ and the mini-batch $\mathcal{B}$, the network structure, and the learning rates of SAC and EISP.
We set the number of subgoals used in Hindsight Sampler as Table \ref{tab:tb_env}.
The weight $\beta$ used in \eqref{eq:loss} refer to Table \ref{tab:tb_env}.
Both the option and action strategies use the Adam optimizer to update the network parameters.
During the training process, we set a time limit of $T=30$ to update the subgoal when the current subgoal is not reached.
For image-based observations, we employ a vector quantized variational autoencoder (VQ-VAE) \cite{van2017neural} to extract features, which are subsequently used as input for subgoal generation. Detailed training procedures and results of the VQ-VAE can be found in Appendix \ref{sec:bootstrapped} and \ref{sec:image}.

\subsection{Qualitative Analysis}
This section demonstrates the feasibility and optimality of inferred subgoals through the qualitative results.

\begin{figure*}[ht]
    \centering
    \includegraphics[width=1.0\textwidth]{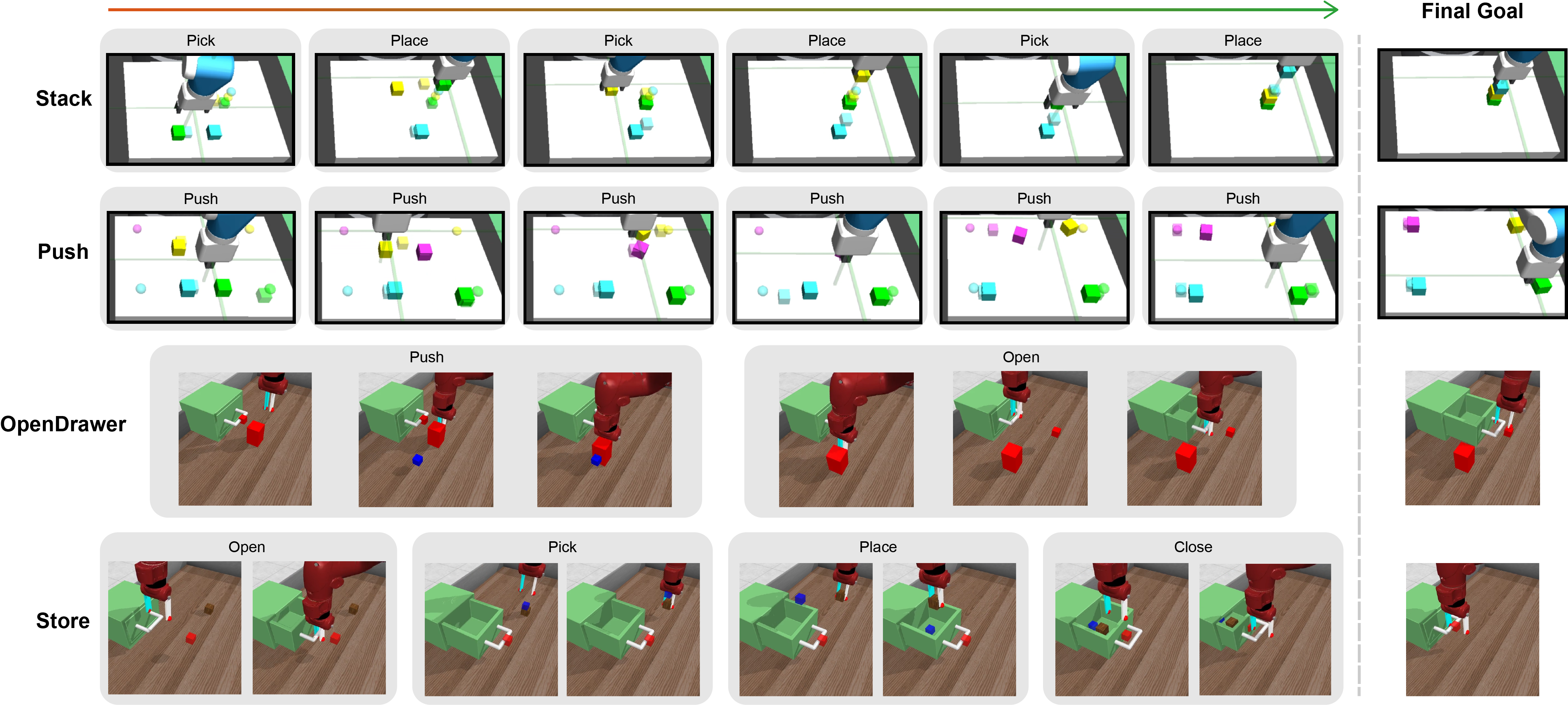}
    \caption{Subgoal sequences generated by option policy $\psi$ from the initial state to desired goal on Stack, Push, OpenDrawer and Store. The transparent cubes in Stack and Push tasks, as well as the red and blue cubes in OpenDrawer and Store tasks, represent subgoals generated by EISP.}
    \label{fig:subgoals-1}
\end{figure*}

\begin{figure}[t]
    \centering
    \includegraphics[width=1.0\columnwidth]{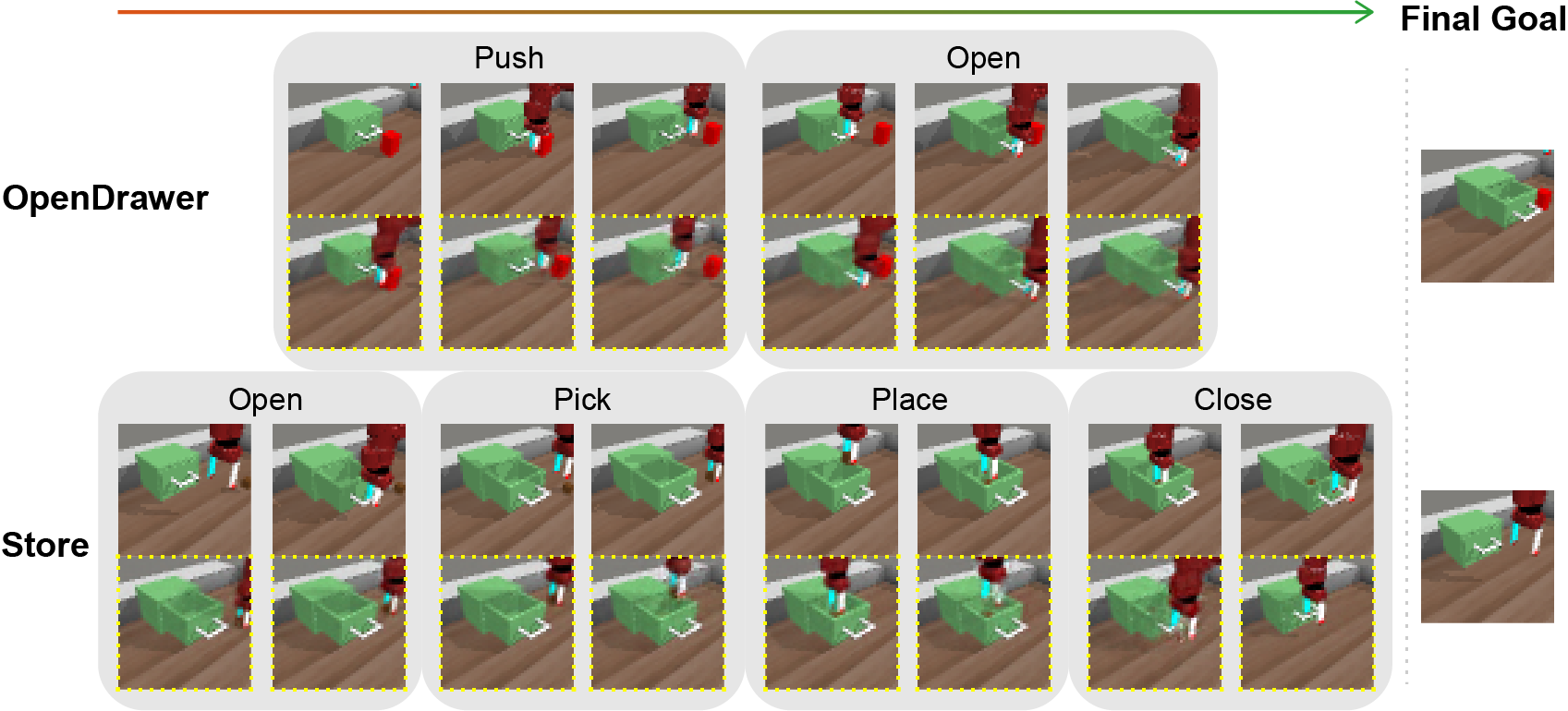}
    \caption{Subgoal sequences generated by option policy $\psi$ from the initial state to desired goal on image-based environments. The figures with yellow dashed rectangles (below) are the subgoals in the current observation (above).}
    \label{fig:subgoals-2}
\end{figure}

\noindent\textbf{Feasibility} Fig. \ref{fig:subgoals-1} shows the subgoals generated by EISP in four state-based tasks mentioned above. We mark the inferred subgoals as transparent cubes in the Stack and Push task and solid small cubes in the OpenDrawer and Store task. For each task, we show the sequence of subgoals in different stages, starting from the initial state to the final goal. Each task involves different skills; for example, the Stack task uses Pick and Place skills iteratively to reach the final goal, whereas the Store task requires four skills to finish the entire task.
Besides, we also infer subgoals for image-based environments, specifically OpenDrawer and Store, with the results shown in Fig. \ref{fig:subgoals-2}. For these tasks, low-dimensional features extracted from image observations serve as input, and the inferred subgoals are reconstructed using a pretrained VQ-VAE decoder. These reconstructed subgoals are depicted in Figure \ref{fig:subgoals-2} with yellow dashed rectangles.

It should be noted that many inferred subgoals are still noisy, indicating that certain subgoals may be unreachable within a given time step. This phenomenon is expected, as these subgoals typically provide the direction to aim for rather than the precise position the robot should achieve. The robot can ultimately achieve the final goal as this direction is progressively updated.
Furthermore, EISP demonstrates robustness against non-optimal or unfeasible subgoals that may emerge during strategy execution, as it infers subgoals based on the current observation.
For instance, the third subgoal generated in the Stack task seems not optimal since it is not directly on the path to the final goal. However, it still provides directional guidance for the robot. As time progresses and the time budget for the current option expires, the next subgoal will offer more accurate information for the robot to achieve its objective.

\begin{figure}[!ht]
    \centering
    \includegraphics[width=0.8\columnwidth]{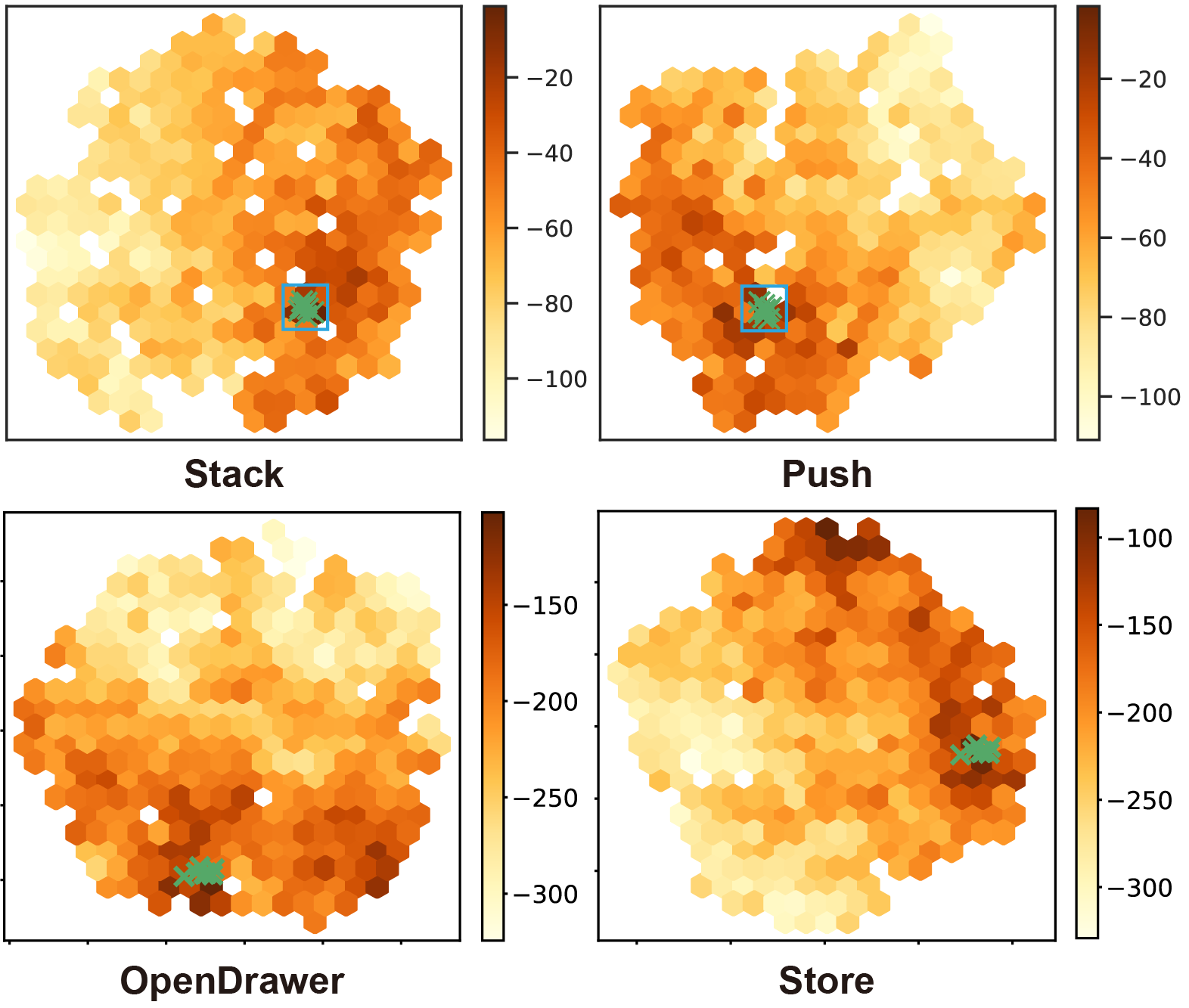}
    \caption{A heatmap of high-dimensional subgoal space visualized by the t-SNE algorithm. The colors are granted according to the magnitude of the $V$ value. Darker colors indicate better subgoals. It is shown that the distributions of the subgoals (marked in green crosses) generated by the option policy share a high $V$ value.}
    \label{fig:heatmap}
\end{figure}

\noindent\textbf{Optimality} To assess the optimality of generated subgoals, we visualize the distribution of subgoals in the high-dimensional subgoal space. For simplicity, the results of later experiments with OpenDrawer and Store were conducted in image-based environments. We compute the $V$ values of 1000 randomly sampled subgoals from the subgoal space of each task. The t-SNE \cite{van2008visualizing} algorithm is then employed to reduce the high-dimensional data into a two-dimensional format. The results are visualized in Fig. \ref{fig:heatmap}. Subgoals with lower $V$ values are marked with lighter hues, indicating they may not be the optimal choice for the current state. In contrast, subgoals with higher $V$ values are shown with darker hues, suggesting they would be more beneficial for the long-horizon task. Subgoal candidates obtained from the trained subgoal strategy are also plotted as green crosses in the figure. The plot reveals that the subgoals inferred by the VAE generator are mainly distributed within the space of higher $V$ values. Initially, these subgoals are not located in regions with high $V$ values, but as the training goes on and the dataset is updated with high-quality data, they are gradually shifted to regions with high $V$ values.

\begin{figure*}[ht]
    \centering
    \includegraphics[width=1.0\textwidth]{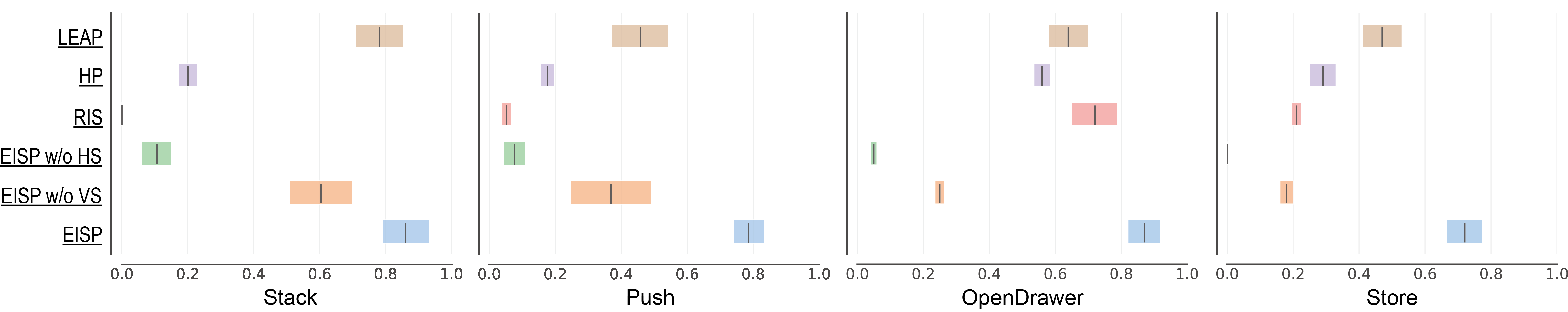}
    \caption{The success rate of EISP and other baselines. The performance of EISP is higher than others in all four environments.}
    \label{fig:success_rate}
\end{figure*}

\subsection{Quantitative Analysis}
Additionally, we perform qualitative experiments to demonstrate the superiority of our proposed methods. We conduct two ablation experiments, with the Hindsight Sampler and the Value Selector components being excluded respectively, to isolate their individual contributions to the overall performance. We also conduct comparative experiments against current state-of-the-art subgoal-generator algorithms, namely \textbf{RIS} (Reinforcement learning with Imaged subgoals) \cite{chane-sane2021goalconditioned}, \textbf{HP} (Hindsight Planner) \cite{lai2020hindsight}, and \textbf{LEAP} (Latent Embeddings for Abstracted Planning) \cite{nasiriany2019planning}.

\begin{itemize}
    \item \textbf{RIS} aims to identify subgoals that are maximally distant from the initial state and final goal.
    \item \textbf{HP} employs the LSTM \cite{hochreiter1997long} architecture to sequentially generate subgoals by continuously integrating previously generated subgoals.
    \item \textbf{LEAP} learns the goal-conditioned policy predicated on the latent embedding of original complex observations.
\end{itemize}

To ensure a fair comparison, all methods employ SAC as the underlying RL algorithm with identical network architecture. The trained policies are tested on four long-horizon tasks, inclusive of our method and the aforementioned baselines, with success rates depicted in Fig. \ref{fig:success_rate}.
The results demonstrate that the EISP algorithm, which utilizes both the Hindsight Sampler and the Value Selector module, yields the highest success rates. The strategy without the Hindsight Sampler, labeled \textbf{EISP w/o HS}, encounters difficulties in rendering the subgoals generated by the RL agent feasible, resulting in low success rates across all four environments. Conversely, the strategy lacking the Value Selector module, denoted as \textbf{EISP w/o VS}, still manages to achieve a measure of success (for instance, approximately a 60\% success rate on the Stack task). This suggests that the optimization module primarily functions as a facilitator to expedite the training of the subgoal generator, enabling the strategy to converge more rapidly and attain a higher success rate.
When compared to other state-of-the-art subgoal-oriented reinforcement learning methods, EISP consistently demonstrates a superior success rate. On the Push task, EISP achieves a success rate exceeding 70\%, while LEAP achieves a success rate of approximately 38\%, and other methods yield success rates of less than 20\%.

\begin{table}[ht]
    \centering
    \caption{Table of the expected returns for different trained policies after testing five times with different seeds. Higher returns indicate better performance.}
    \begin{tabular}{ccccc}
        \toprule
        Policy               & Stack                 & Push                  & OpenDrawer           & Store               \\
        \midrule
        \textbf{RIS}         & $-269.46$             & $-270.27$             & \underline{-256.374} & $-865.453$          \\
        \textbf{HP}          & $-283.46$             & $-311.96$             & $-272.97$            & \underline{-844.58} \\
        \textbf{LEAP}        & \underline{$-159.80$} & \underline{$-245.29$} & $-266.34$            & $-639.97$           \\
        \textbf{EISP w/o HS} & $-284.67$             & $-344.58$             & $-283.75$            & $-990.43$           \\
        \textbf{EISP w/o VS} & $-201.33$             & $-313.15$             & $-284.72$            & $-775.61$           \\
        \textbf{EISP (Ours)} & \textbf{-89.29}       & \textbf{-228.89}      & \textbf{-246.91}     & \textbf{-581.96}    \\
        \bottomrule
    \end{tabular}
    \label{tab:return}
\end{table}
After conducting five tests on four tasks using distinct seeds, the expected returns recorded for all the training policies are presented in Table \ref{tab:return}, with the highest return for each task emphasized in bold and the second highest return emphasized in underlining.
A higher expected return indicates a superior policy. While all policies can potentially lead to higher expected returns, our approach produces the most substantial improvements in expected returns.
The LEAP and HP methods rely solely on the initial state to infer subgoals, which proves insufficient for tasks with varying desired goals. These methods lack robustness for complex long-horizon tasks in most scenarios, and a change in the desired goal results in the generated subgoals becoming less accurate, leading to task failure.
Furthermore, sampling subgoals randomly from the experience buffer can result in inaccurate subgoals, impeding the generation of optimal lower-level action policies. For detailed information on episodic returns throughout the entire training procedure, you can refer to Appendix \ref{sec:return}.

\begin{figure}[t]
    \centering
    \includegraphics[width=\columnwidth]{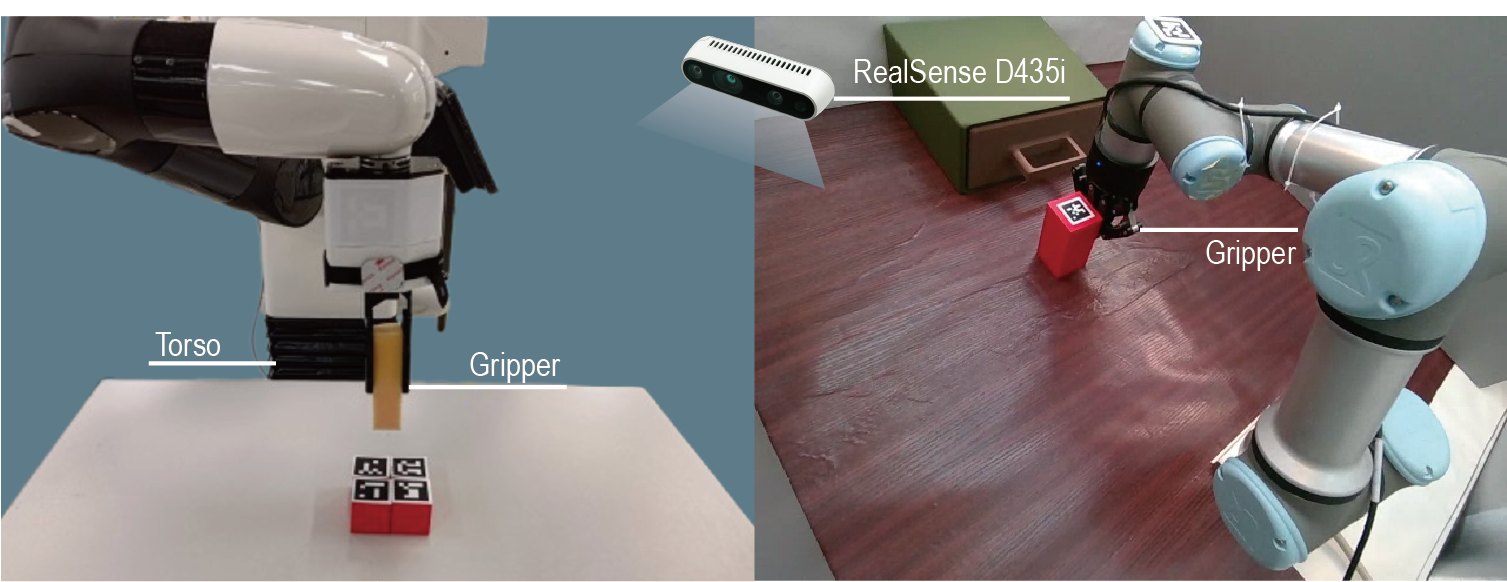}
    \caption{Environments setup. Left: Tiago robotic arm. Right: UR3 robotic arm.}
    \label{fig:env_real}
\end{figure}

\subsection{Real World Demonstrations}

\begin{figure}[t]
    \centering
    \includegraphics[width=\columnwidth]{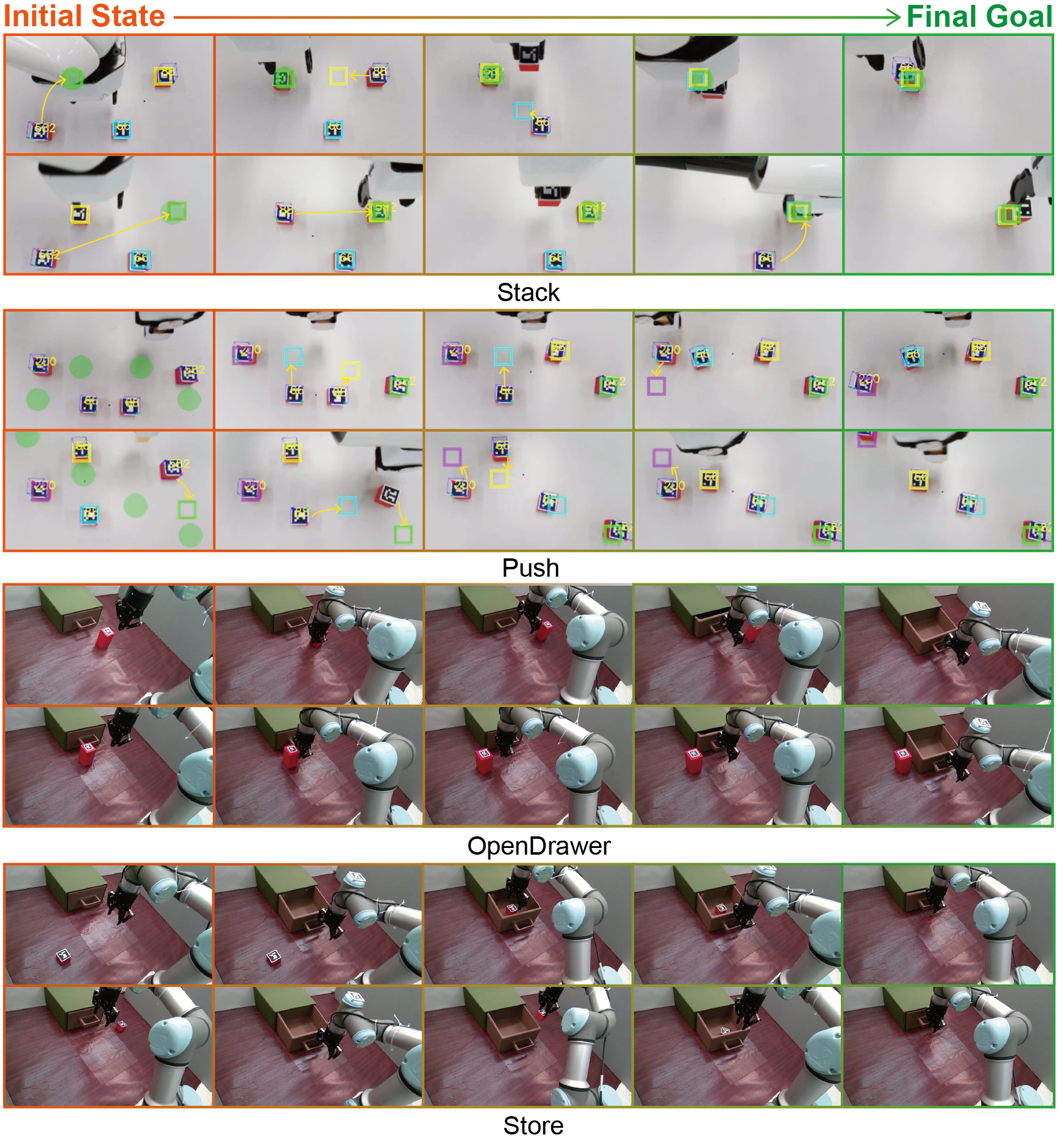}
    \caption{The trajectories generated by the four manipulation tasks when executed in a real environment are presented. For the Stack and Push tasks, green spherical shadows represent the final goals, and transparent squares denote the subgoals generated during execution.}
    \label{fig:real_exp}
\end{figure}

To demonstrate the adaptability of our algorithms in the real world, we employ the Tiago++ robotic arm \cite{pages2016tiago} to execute Stack and Push tasks, and the UR3 robotic arm for OpenDrawer and Store tasks.
The setups of these four environments are the same as in the simulation.
Our experiments utilize Pinocchio \cite{carpentier2019pinocchio} for both motion planning and inverse kinematics. As depicted in Fig. \ref{fig:env_real}, red cubes are manipulated by the robotic arm to complete the assigned task, with the number of red blocks used in Stack and Push being 3 and 4, respectively. We mount a calibrated Intel RealSense D435i RGB-D camera on the top of the table to facilitate top-view observation. We also fix a soft beam to the end of the gripper to mitigate potential collisions between the gripper, the table, and the blocks during the execution of the Push task.

The strategy employed to guide the robot is identical to that used in the simulation. To gather environmental observations, we utilize Aruco markers for the detection of each object's and gripper's position. Fig. \ref{fig:real_exp} illustrates the subgoals inferred from the initial state to the desired goals for four manipulation tasks. The desired goals are denoted as green spherical shades and the subgoals as transparent squares.
During execution, the Aruco markers may be occluded by the robotic arm, causing the top-view camera unable to detect the markers at certain time steps, primarily in the Stack and Push task. To address this problem, we maintain the last known position of the markers and update it when markers become detectable again. Additionally, the robot arm is reset to a specific pose to observe the current state after achieving each subgoal. We also conducted robustness tests and failed case studies, with the results presented in Appendix \ref{sec:real} and \ref{sec:failed}.

\section{Conclusion}\label{sec:conclusion}
In conclusion, we propose Explicit-Implicit Subgoal Planning (EISP), a novel algorithm that leverages an explicit encoder model to produce feasible subgoals and an implicit decoder model that provides a guarantee on the worst-case log-likelihood of the subgoal distribution.
The efficiency of the proposed algorithm is substantiated by both qualitative and quantitative results derived from simulations and real-world experiments, demonstrating its superiority over prevailing baselines in terms of subgoal efficacy and overall performance.

As we learned from the failure analysis, the generated subgoals are not conducive to long-horizon task accomplishment under extreme distributions, i.e., too far or too close to the current state.
Future research will delve deeper into integrating with the exploration method to ensure that the subgoals are within a reasonable range from the current achieved goals, which can make the generated subgoals more accurate. Despite these limitations, the approach presented in this paper offers a promising avenue for designing effective divide-and-conquer strategies through abstract subgoal reasoning in complex robotics tasks.

\ifCLASSOPTIONcaptionsoff
    \newpage
\fi

\bibliographystyle{IEEEtran}
\bibliography{main}

\begin{IEEEbiography}
    [{\includegraphics[width=1in,height=1.25in,clip,keepaspectratio]{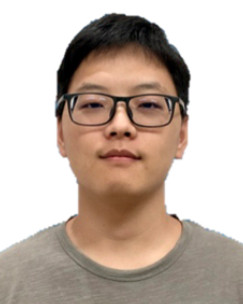}}]
    {Fangyuan Wang} received the M.Sc. degree in software engineering from Zhejiang Sci-Tech University, China, in 2022.
    He is currently pursuing the Ph.D. in mechanical engineering at The Hong Kong Polytechnic University, Hong Kong.
    His research interests focus on reinforcement learning, multi-agent systems, and robotic manipulation.
\end{IEEEbiography}

\begin{IEEEbiography} [{\includegraphics[width=1in,height=1.25in,clip,keepaspectratio]{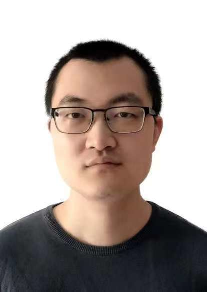}}]
    {Anqing Duan} received his bachelor's degree in mechanical engineering from Harbin Institute of Technology, Harbin, China, in 2015, his master's degree in mechatronics from KTH, Sweden, in 2017, and his Ph.D. degree in robotics from the  Italian Institute of Technology and the University of Genoa, Italy, in 2021.
    Since 2021, he has been a Research Associate with The Hong Kong Polytechnic University.
    His research interest includes robot learning and control.
\end{IEEEbiography}

\begin{IEEEbiography}
    [{\includegraphics[width=1in,height=1.25in,clip,keepaspectratio]{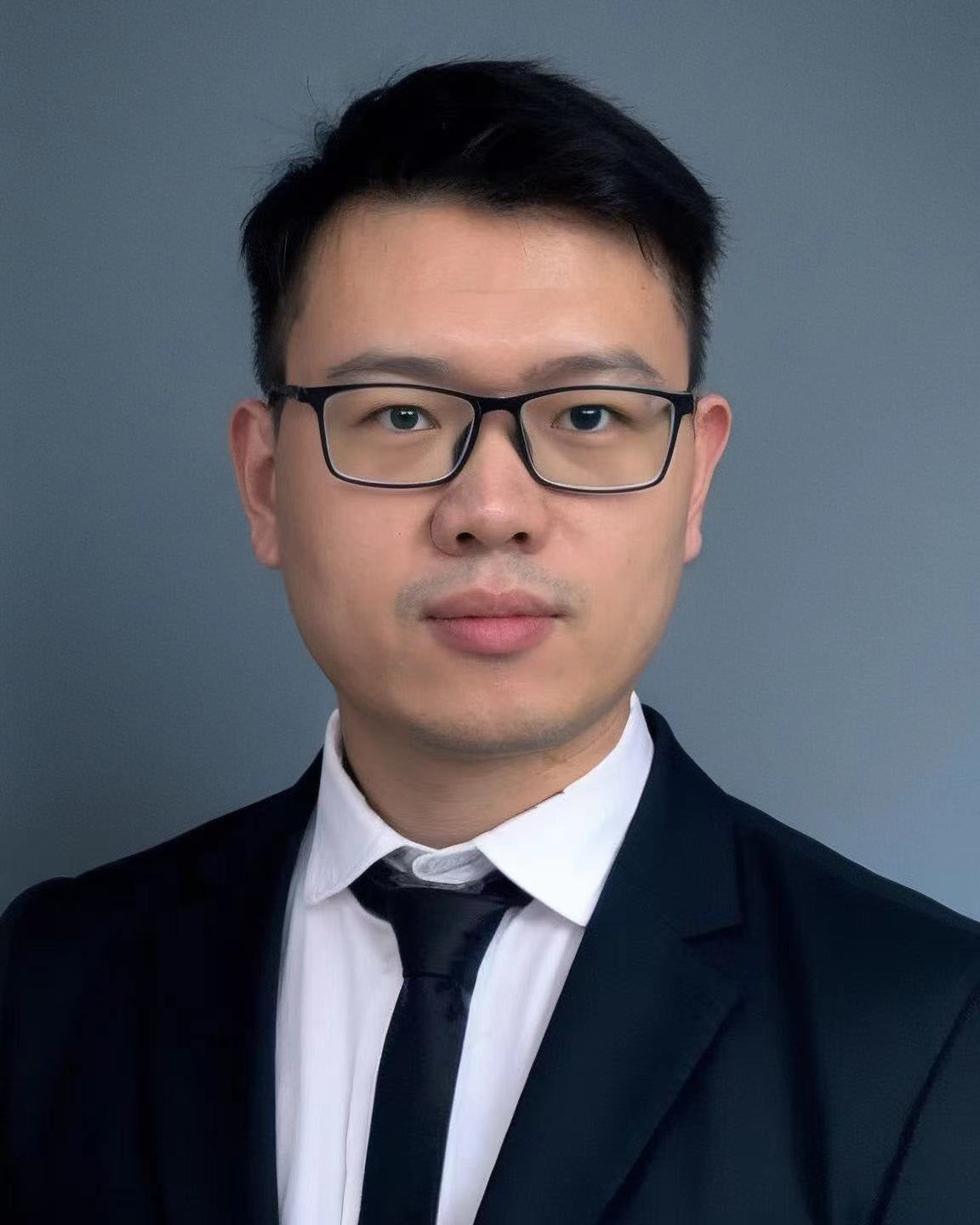}}]
    {Peng Zhou} received his Ph.D. degree in robotics from PolyU, Hong Kong, in 2022.
    In 2021, he was a visiting Ph.D. student at KTH Royal Institute of Technology, Stockholm, Sweden.
    He is currently a Research Officer at the Centre for Transformative Garment Production and a Postdoctoral Research Fellow at The University of Hong Kong. His research interests include deformable object manipulation, robot reasoning and learning, and task and motion planning.
\end{IEEEbiography}

\begin{IEEEbiography}
    [{\includegraphics[width=1in,height=1.25in,clip,keepaspectratio]{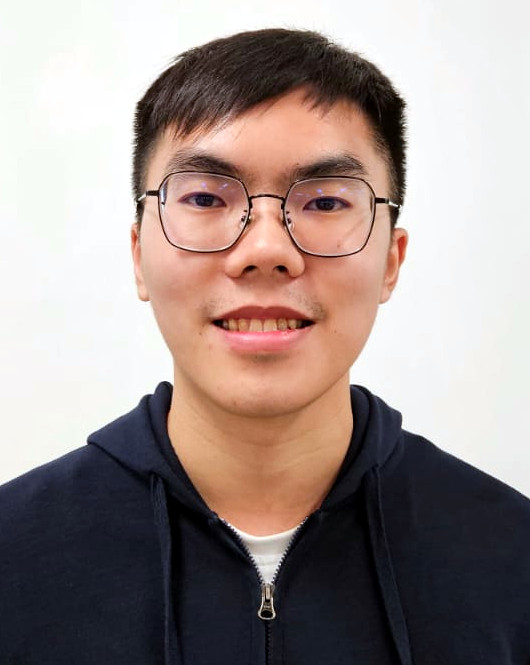}}]
    {Shengzeng Huo} received the B.Eng. degree in vehicle engineering from the South China University of Technology, China, in 2019, and the M.Sc. degree in mechanical engineering from The Hong Kong Polytechnic University, Hong Kong, in 2020, and he is currently pursuing the Ph.D. degree in the same discipline since 2021. His current research interests include bimanual manipulation, deformable object manipulation, and robot learning.
\end{IEEEbiography}

\begin{IEEEbiography}
    [{\includegraphics[width=1in,height=1.25in,clip,keepaspectratio]{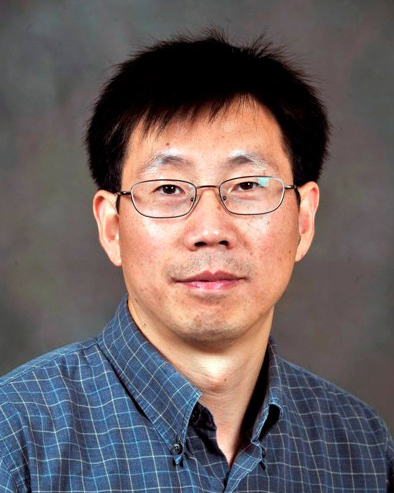}}]
    {Guodong Guo} (M'07-SM'07) received the B.E. degree in Automation
    from Tsinghua University, Beijing, China,
    the Ph.D. degree in Computer Science from the University of Wisconsin, Madison, WI, USA.
    He is now a Professor at Eastern Institute of Technology,
    and the Vice President of Ningbo Institute of Digital Twin, China.
    He is also affiliated with the Dept. of Computer Science and Electrical Engineering,
    West Virginia University, USA.
    His research interests include computer vision, biometrics, machine learning, and
    multimedia. He is an AE of several journals, including IEEE Trans. on Affective Computing.
    He received the North Carolina State Award for Excellence in Innovation in 2008,
    New Researcher of the Year (2010-2011),
    and Outstanding Researcher (2017-2018, 2013-2014) at CEMR, WVU.
    He was selected the “People’s Hero of the Week” by BSJB under Minority
    Media and Telecommunications Council (MMTC) in 2013. His papers
    were selected as “The Best of FG’13” and “The Best of FG’15”, respectively,
    and the “Best Paper Award” by the IEEE Biometrics Council in 2022.
\end{IEEEbiography}

\begin{IEEEbiography}
    [{\includegraphics[width=1in,height=1.25in,clip,keepaspectratio]{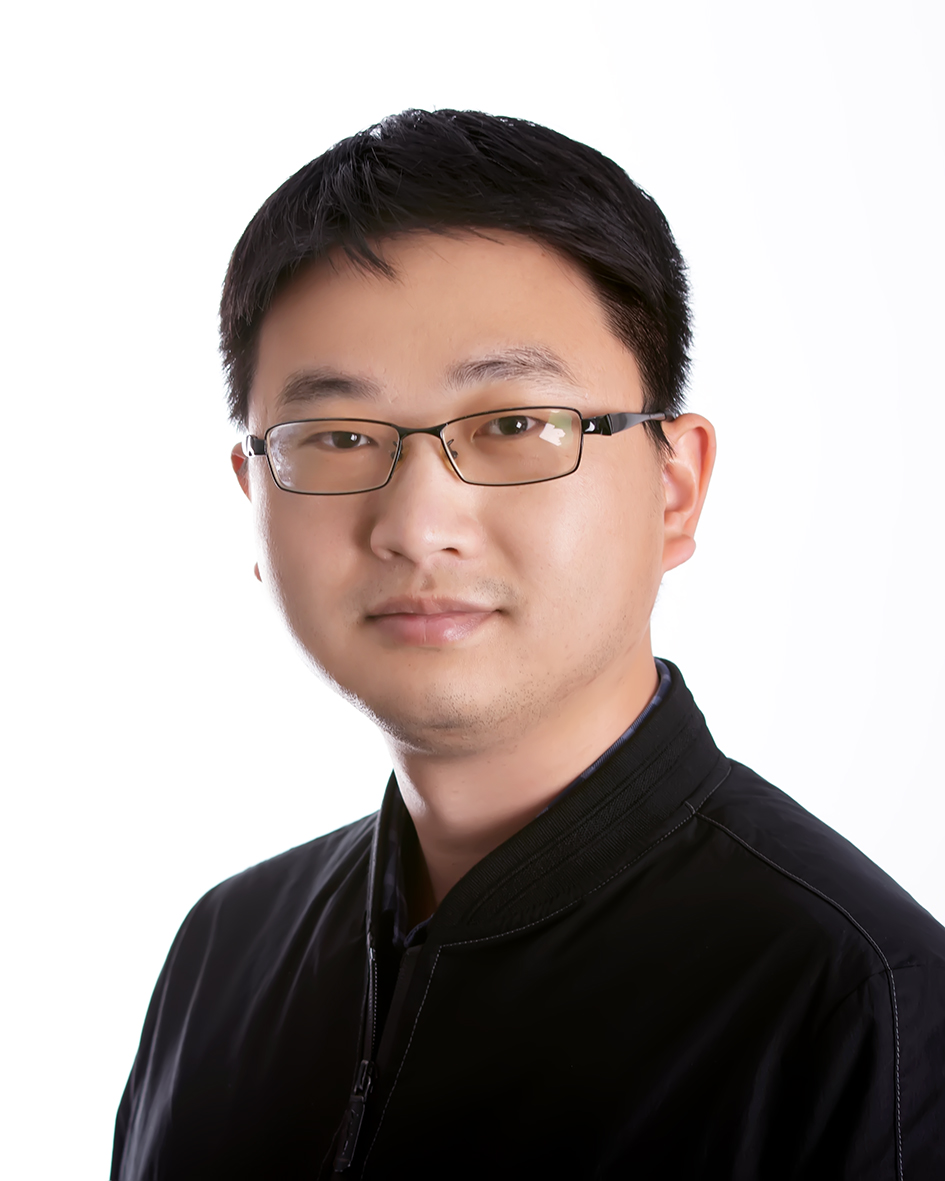}}] {Chenguang Yang} (Fellow, IEEE) received the B.Eng. degree in measurement and control from Northwestern Polytechnical University, Xian, China, in 2005, and the Ph.D. degree in control engineering from the National University of Singapore, Singapore, in 2010. He performed postdoctoral studies in human robotics at the Imperial College London, London, U.K from 2009 to 2010.  He is Chair in Robotics with Department of Computer Science, University of Liverpool, UK.  He was awarded UK EPSRC UKRI Innovation Fellowship and individual EU Marie Curie International Incoming Fellowship. As the lead author, he won the IEEE Transactions on Robotics Best Paper Award (2012) and IEEE Transactions on Neural Networks and Learning Systems Outstanding Paper Award (2022). He is the Corresponding Co-Chair of IEEE Technical Committee on Collaborative Automation for Flexible Manufacturing. His research interest lies in human robot interaction and intelligent system design.
\end{IEEEbiography}

\begin{IEEEbiography}
    [{\includegraphics[width=1in,height=1.25in,clip,keepaspectratio]{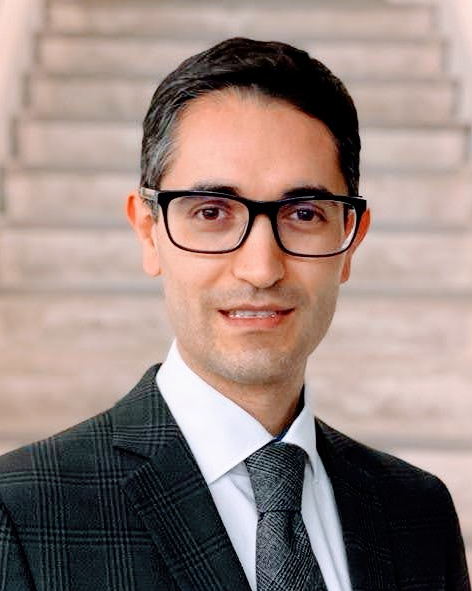}}] {David Navarro-Alarcon} (Senior Member, IEEE) received his Ph.D. degree in mechanical and automation engineering from The Chinese University of Hong Kong in 2014.
    He is currently an Associate Professor at the Department of Mechanical Engineering at The Hong Kong Polytechnic University (PolyU). His current research interests include perceptual robotics and control theory. He currently serves as an Associate Editor of the \textsc{IEEE Transactions on Robotics}.
\end{IEEEbiography}

\onecolumn
\appendix

\subsection{Probability of the feasibility} \label{sec:feasibility}

To ensure a given subgoal sequence $\hat{g}_1, \hat{g}_2, \cdots, \hat{g}_{n-1}$ is feasible, we must guarantee that the transition probability $\rho(\hat{g}_{i-1}, \hat{g}_{i})$ from current subgoal $\hat{g}_{i-1}$ to next subgoal $\hat{g}_{i}$ exceeds 0.

Assuming the initial state of the current subgoal is feasible to reach, i.e., the probability of $\rho(s=s_0^i) > 0$. The transition probability can be obtained as follows:

\begin{align*} \label{eq:fp}
    \begin{split}
        \rho(\hat{g}_{i-1}, \hat{g}_{i}) & = \rho(\phi(s_0^i), \phi(s_e^i))                                                                                \\
                                         & = \rho(s_0^i, s_0^e)                                                                                            \\
                                         & = \rho(s=s_0^i) \prod_{s = s_{0}^i, a \sim \pi_i}^{s_e^i} \left[ \pi_{i}(s, \hat{g}_i)\mathcal{T}(s, a) \right]
    \end{split}
\end{align*}

\subsection{Derivation of the objective of the Hybrid Subgoal Generator} \label{sec:derivation}
\begin{proof}
    Given the initial state $s_0^i$ of option $i$, we want to maximize the log probability of the final goal $\log{p_{\psi^{\prime}}(g)}$.

    \begin{align*}
        \log p_{\psi^{\prime}}(g) & = \log p_{\psi^{\prime}}(g) \int p_{\psi}(\hat{g}_i|g,s_0^i) \hat{d\hat{g}_i} \tag{$\int p_{\psi}(\hat{g}_i|g,s_0^i) = 1$}                                                                                                                                             \\
                                  & = \int \log p_{\psi^{\prime}}(g) p_{\psi}(\hat{g}_i|g,s_0^i)d\hat{g}_i                                                                                                                                                                                                 \\
                                  & = \mathbb{E}_{\psi(\hat{g}_i|g,s_0^i)} \Bigg[\log p_{\psi^{\prime}}(g)\Bigg]                                                                                                                                                                                           \\
                                  & = \mathbb{E}_{\psi(\hat{g}_i|g,s_0^i)} \Bigg[ \log \frac{p_{\psi^{\prime}}(\hat{g}_i,g,s_0^i)}{p_{\psi^{\prime}}(\hat{g}_i|g,s_0^i) p_{\psi^{\prime}}(s_0^i|g,\hat{g}_i)} \Bigg] \tag{Chain rule}                                                                      \\
                                  & = \mathbb{E}_{\psi(\hat{g}_i|g,s_0^i)} \Bigg[ \log \frac{p_{\psi^{\prime}}(\hat{g}_i,g,s_0^i)}{p_{\psi^{\prime}}(\hat{g}_i|g,s_0^i) } \Bigg] \tag{$p_{\psi}(s_0^i|g,\hat{g}_i) = 1$}                                                                                   \\
                                  & = \mathbb{E}_{\psi(\hat{g}_i|g,s_0^i)} \Bigg[ \log \frac{p_{\psi^{\prime}}(\hat{g}_i,g,s_0^i)}{p_{\psi^{\prime}}(\hat{g}_i|g,s_0^i)} \frac{p_{\psi}(\hat{g}_i|g,s_0^i)}{p_{\psi}(\hat{g}_i|g,s_0^i)} \Bigg]                                                            \\
                                  & = \mathbb{E}_{\psi(\hat{g}_i|g,s_0^i)} \Bigg[ \log \frac{p_{\psi^{\prime}}(\hat{g}_i,g,s_0^i)}{p_{\psi}(\hat{g}_i|g,s_0^i)} \Bigg]  + \mathbb{E}_{\psi(\hat{g}_i|g,s_0^i)} \Bigg[ \log \frac{p_{\psi}(\hat{g}_i|g,s_0^i)}{p_{\psi^{\prime}}(\hat{g}_i,g,s_0^i)} \Bigg] \\
    \end{align*}

    where $\mathbb{E}_{p_{\psi}(\hat{g}_i|g,s_0^i)}[\log\frac{p_{\psi^{\prime}(g,\hat{g}_i, s_0^i)}}{p_{\psi}(\hat{g}_i|g,s_0^i)}]$ is the evidence lower bound (\textbf{ELBO}) and \linebreak $\mathbb{E}_{\psi(\hat{g}_i|g,s_0^i)} [ \log \frac{p_{\psi}(\hat{g}_i|g,s_0^i)}{p_{\psi^{\prime}}(\hat{g}_i,g,s_0^i)} ]$ is the KL divergence greater than 0.
    Thus, the objective now is:

    \begin{align*}
        \log p_{\psi^{\prime}}(g) & \geq \textbf{ELBO}                                                                                                                                                                                                                                            \\
                                  & = \mathbb{E}_{p_{\psi}(\hat{g}_i|g,s_0^i)} \Bigg[\log\frac{p_{\psi^{\prime}}(g,\hat{g}_i, s_0^i)}{p_{\psi}(\hat{g}_i|g,s_0^i)} \Bigg]                                                                                                                         \\
                                  & = \mathbb{E}_{p_{\psi}(\hat{g}_i|g,s_0^i)} \Bigg[\log\frac{p_{\psi^{\prime}}(g | \hat{g}_i, s_0^i) p_{\psi^{\prime}}(s_0^i, \hat{g}_i)}{p_{\psi}(\hat{g}_i|g,s_0^i)} \Bigg]                                                                                   \\
                                  & = \mathbb{E}_{p_{\psi}(\hat{g}_i|g,s_0^i)} \Bigg[ \log p_{\psi^{\prime}} (g | \hat{g}_i, s_0^i) \Bigg] + \mathbb{E}_{p_{\psi}(\hat{g}_i|g,s_0^i)} \Bigg[ \log \frac{p_{\psi^{\prime}}(s_0^i, \hat{g}_i)}{p_{\psi}(\hat{g}_i|g,s_0^i)} \Bigg] \tag{Chain rule} \\
                                  & = \mathbb{E}_{p_{\psi}(\hat{g}_i|g,s_0^i)} \Bigg[ \log p_{\psi^{\prime}} (g | \hat{g}_i, s_0^i) \Bigg] - D_{KL}\Bigg[p_{\psi}(\hat{g}_i|s_0^i, g)\|p(\hat{g}_i)\Bigg] \tag{$p_{\psi^{\prime}}(\hat{g}_i,s_0^i) = p_{\psi^{\prime}}(\hat{g}_i)$}
    \end{align*}
\end{proof}

\subsection{Bootstrapped data collection} \label{sec:bootstrapped}
Deep reinforcement learning not only allows us to extract the best strategies from bootstrapped data but also allows robots to continually improve as they increasingly interact with spoiled data.
To accelerate training, we collect demo trajectories for offline policy training using a scripted policy that utilizes privileged information from the environment (e.g., object pose and fixture state).
We collect $3000$ trajectories for offline policy training using scripted policies that utilize privileged information from the environment (e.g., object pose and fixture state).
Within each demo trajectory, we initialize the environment randomly and set an arbitrary level of the desired goal. Trajectory lengths range from $100$ to $400$ for different tasks.
The success rate of the scripted policies is no greater than $10\%$.

\subsection{Image-based observation processing} \label{sec:image}
For the OpenDrawer and StoreBlock tasks, we also use image-based observation to train the subgoal generator. The observation is a $3 \times 48 \times 48$ image, which is processed by a vector quantized variational autoencoder (VQ-VAE) \cite{van2017neural} first to extract the features \cite{khazatsky2021can}. To elaborate, we pretrain an encoder, denoted as $\phi$, which takes the original image as input and outputs the latent embedding of the image. The resulting low-dimensional latent representation is then input to the subgoal generator. The VQ-VAE comprises a convolutional encoder and a deconvolutional decoder. The encoder consists of three convolutional layers with kernel size $4 \times 4$ and stride $2 \times 2$. The decoder consists of three deconvolutional layers with kernel size $4 \times 4$ and stride $2 \times 2$. The latent space size is $3 \times 6 \times 6$. The VQ-VAE is trained using the Adam optimizer with a learning rate of $1e-4$ and a batch size of $256$ for $1e6$ steps. Then, the subgoal generator for the image-based environment is trained using the Adam optimizer with a learning rate of $3e-5$ and a batch size of $1024$.

\begin{figure}[ht]
    \centering
    \includegraphics[width=0.8\columnwidth]{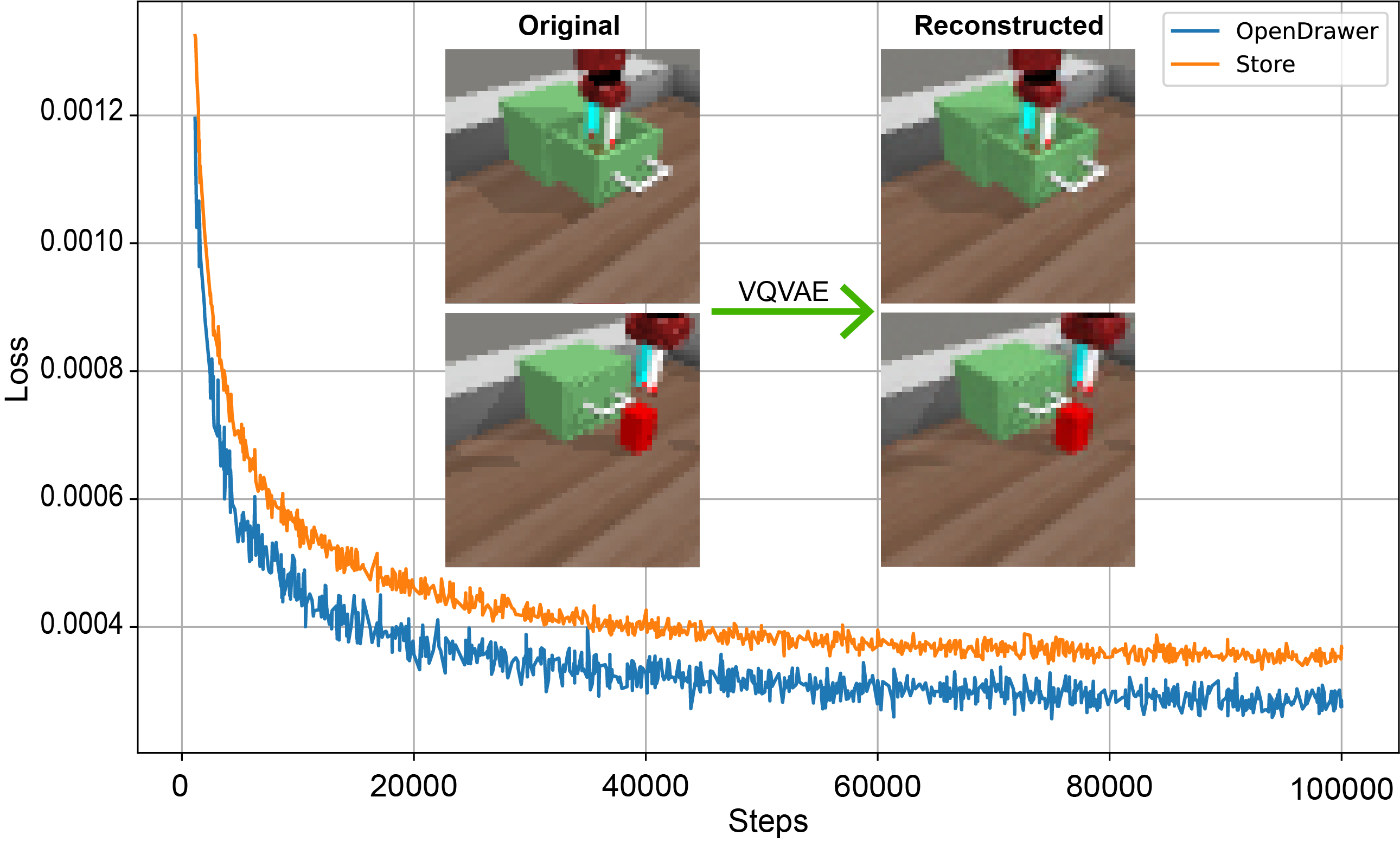}
    \caption{The VQ-VAE loss during OpenDrawer and Store tasks training. The upper and lower images show the original and reconstructed images, respectively.}
    \label{fig:vqvae-loss}
\end{figure}

Fig.\ref{fig:vqvae-loss} shows the training process of the VQ-VAE on the OpenDrawer and Store tasks. The VQ-VAE loss progressively decreases as the training progresses. Additionally, we present two examples comparing the original and reconstructed images. The upper images depict the original images, while the lower images show the reconstructed counterparts. The reconstructed images closely resemble the original ones, indicating that the VQVAE effectively extracts features from the image observations and reconstructs them using embeddings.
We utilize the features extracted by the VQ-VAE, which are low-dimensional data (around 8 times smaller than the original), which are then fed into the subgoal generator to generate the subgoals.

\newpage
\subsection{Episodic return}\label{sec:return}

\begin{figure*}[h!]
    \centering
    \includegraphics[width=0.8\columnwidth]{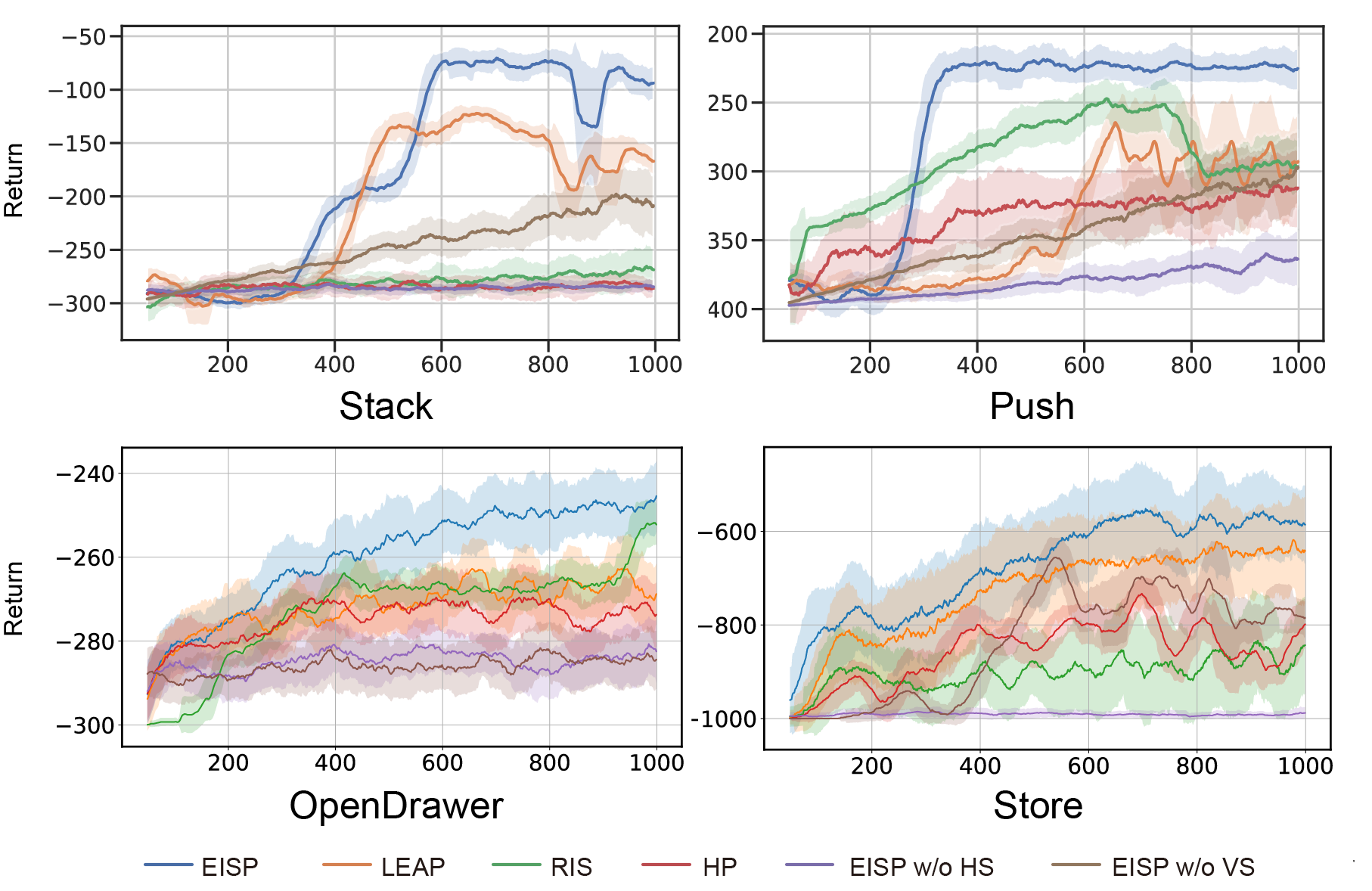}
    \caption{The episodic expected return during training. EISP can achieve the highest expected return among all baseline methods.}
    \label{fig:return}
\end{figure*}

The evolution of the expected reward during training, i.e., the cumulative reward over the entire trajectory, is depicted in Fig. \ref{fig:return}.

\subsection{Real world tasks analysis}\label{sec:real}

\begin{figure}[t]
    \centering
    \includegraphics[width=0.6\columnwidth]{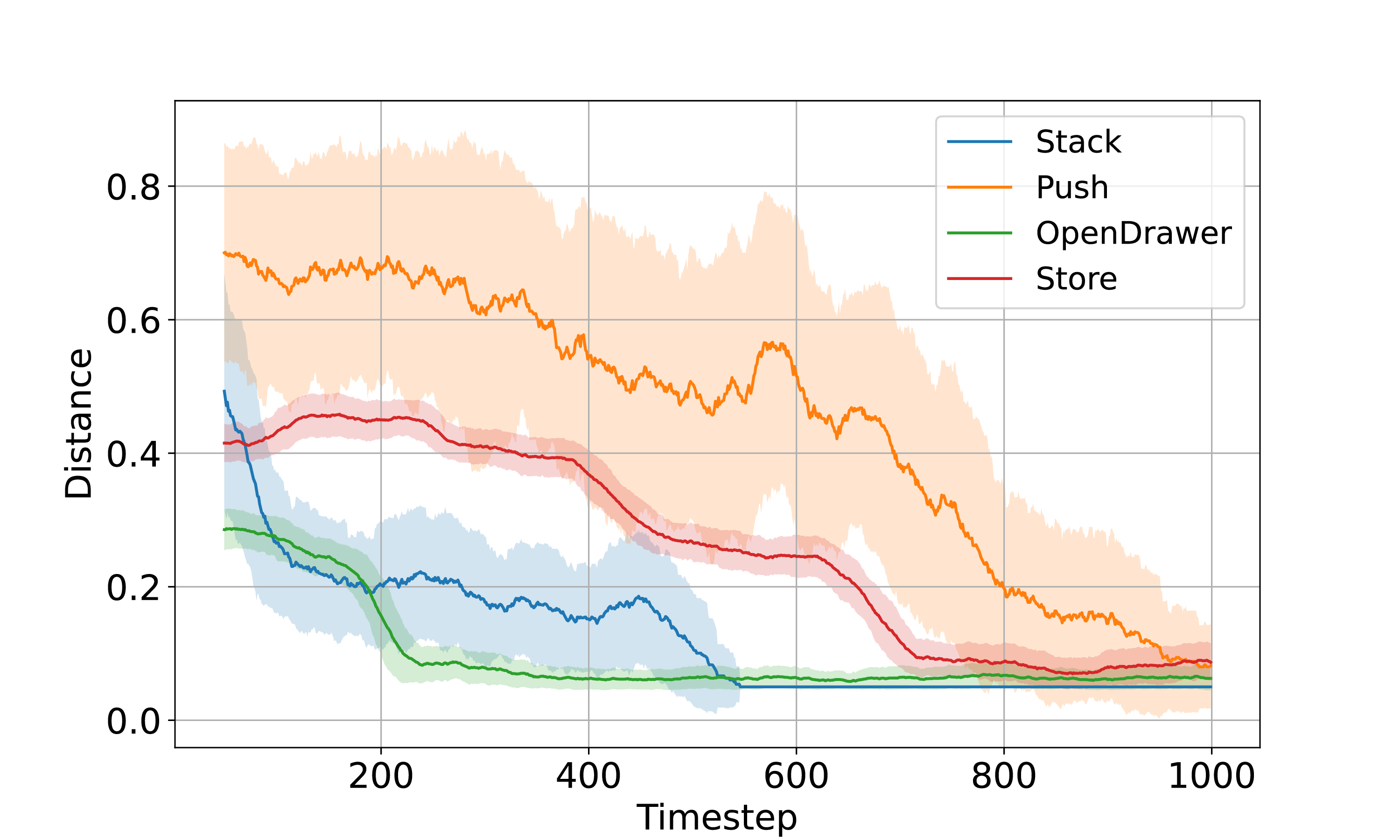}
    \caption{The distance to the desired goal from the current state when executing in the real world is measured.}
    \label{fig:distance}
\end{figure}

Fig. \ref{fig:distance} illustrates the evolution of the distance between the current achieved goal and the final goal over the task execution. The distance is calculated as the average of the Euclidean distances between the positions of the manipulating objects and their corresponding target positions. A task is deemed successful if this distance falls within a threshold of $0.05m$. As the experiment goes on, the distance for all tasks steadily declines to 0.05m.

\begin{table}[h]
    \centering
    \caption{The success rate of three manipulation tasks in the real world.}
    \begin{tabular}{ccc}
        \toprule
        Task                          & Stack & Push4 \\
        \midrule
        No interference               & 8/10  & 8/10  \\
        Interference after completion & 8/10  & 10/10 \\
        \bottomrule
    \end{tabular}
    \label{tab:success_real}
\end{table}

\begin{figure}[h!]
    \centering
    \includegraphics[width=\columnwidth]{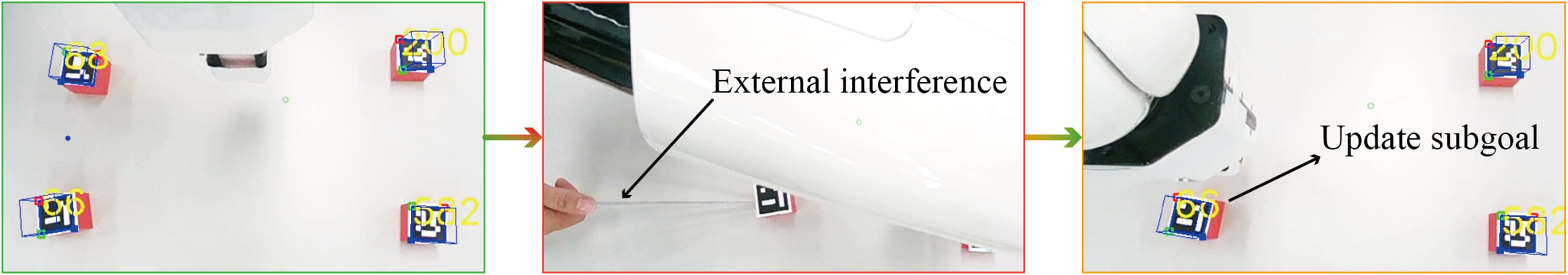}
    \caption{The performance of the algorithm when the current state is externally disturbed by humans. The robot will update the subgoals in real-time.}
    \label{fig:robust}
\end{figure}

We also conducted ten tests for the Stack and Push task, incorporating human disturbances to interfere with the completed task. The success rate of each task is detailed in Table \ref{tab:success_real}. The results indicate that EISP achieves higher success rates of 80\% and 100\% for the Stack and Push tasks, respectively, despite human interaction.
As depicted in Fig. \ref{fig:robust}, when a block is moved away from its designated goal intentionally, the subgoal is promptly updated upon detection of the block, thereby guiding the gripper to complete the subtask.
EISP possesses the capacity to infer subgoals from arbitrary states due to its sequential planning characteristics. In contrast, other subgoal generators, such as LEAP and HP, lack robustness because they infer subgoals only at the initial stage.

\subsection{Failed cases analysis}
\label{sec:failed}

\begin{figure}[h!]
    \centering
    \includegraphics[width=0.6\columnwidth]{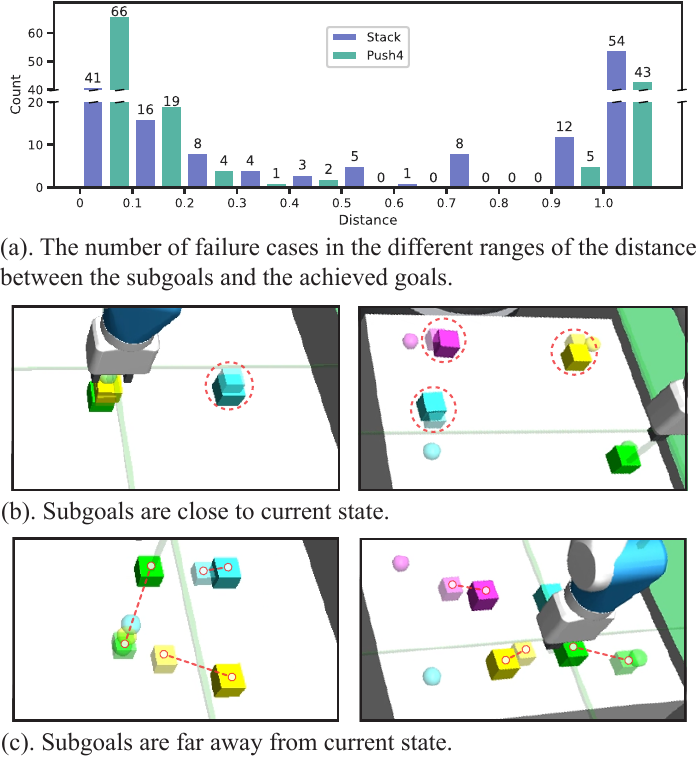}
    \caption{The analysis of failed cases reveals that when subgoals are generated either too close to or too far from the current state, it becomes challenging for the agent to achieve them.}
    \label{fig:failed}
\end{figure}

Although our algorithm generally performs well, occasional failures are observed, particularly in the Stack and Push4 tasks. We conduct 1000 tests on each of these tasks on different initial states and final goals, and quantify the failures across varying ranges of distance between the generated subgoal and the goal achieved at the end of the episodes.
As shown in Fig. \ref{fig:failed} (a), a notably high number of failures in the distance range of $0$ to $0.1$ and above $1.0$, far surpassing the failure count in intermediate distance ranges.
We analyze that the following two reasons may cause it. First, the subgoal is extremely close to the current state, as shown in Fig. \ref{fig:failed} (b), indicating that the action policy prematurely deems the subgoal as achieved due to its minimal difference from the current state.
Secondly, the subgoal is significantly far from the current state, as depicted in Fig. \ref{fig:failed} (c), implying that the generated subgoal is still considered a long-horizon task, challenging the action policy to accomplish.

\end{document}